\documentclass[10pt,twocolumn,letterpaper]{article}

\usepackage[pagenumbers]{3dv/cvpr} %

\makeatletter
\@namedef{ver@everyshi.sty}{2001/05/15}
\makeatother

\definecolor{cvprblue}{rgb}{0.21,0.49,0.74}
\usepackage[pagebackref,breaklinks,colorlinks,allcolors=cvprblue]{hyperref}

\def\paperID{83} %
\def\confName{3DV\xspace}
\def\confYear{2027\xspace}

\title{\method{}: Generative Reconstruction with Test-Time Input Alignment}

\author{
Stefano Esposito${}^1$, Naama Pearl${}^1$, Polina Karpikova${}^1$,
Samuel Rota Bul\`o${}^2$,\\
Lorenzo Porzi${}^2$, Peter Kontschieder${}^2$,
Andreas Geiger${}^1$, Jonathon Luiten${}^2$\\
\vspace{2pt}\\
${}^1$T\"ubingen AI Center, University of T\"ubingen \qquad 
${}^2$Meta Reality Labs
}

\usepackage[utf8]{inputenc} %
\usepackage[T1]{fontenc}    %
\usepackage{textcomp}       %

\providecommand{\failmark}{\ensuremath{\times}}
\providecommand{\warnmark}{\textsuperscript{\ensuremath{\triangle}}}
\DeclareUnicodeCharacter{274C}{\failmark}  %
\DeclareUnicodeCharacter{26A0}{\warnmark}  %
\DeclareUnicodeCharacter{FE0F}{}           %

\usepackage{hyperref}       %
\usepackage{url}            %
\usepackage{booktabs}       %
\usepackage{amsfonts}       %
\usepackage{nicefrac}       %
\usepackage{microtype}      %
\PassOptionsToPackage{table}{xcolor}
\usepackage{xcolor}         %

\usepackage{amsmath,amsfonts,amssymb,amsthm}
\usepackage{mathtools}  %
\usepackage{caption}
\usepackage{subcaption}

\usepackage{graphicx}
\graphicspath{{./}{../data/}{figures/}}
\usepackage{lipsum}
\usepackage{multirow}   %
\usepackage{makecell}   %
\usepackage{colortbl}   %
\usepackage{enumitem}
\usepackage{adjustbox}

\usepackage{pgfplots}
\pgfplotsset{compat=1.18}

\newcommand{\bc}{\mathbf{c}}

\newcommand{\bt}{\mathbf{t}}

\newcommand{\bv}{\mathbf{v}}

\newcommand{\bx}{\mathbf{x}}

\DeclareMathOperator*{\argmin}{argmin~}

\makeatletter
\DeclareRobustCommand\onedot{\futurelet\@let@token\@onedot}
\def\@onedot{\ifx\@let@token.\else.\null\fi\xspace}

\makeatother

\newcommand{\boldparagraph}[1]{\vspace{0.2cm}\noindent{\bf #1:} }

\definecolor{darkgreen}{rgb}{0,0.7,0}

\usepackage{tikz}
\usetikzlibrary{positioning,calc}

\newcommand{\qbadgefontstyle}{\rmfamily}  %
\newcommand{\qbadgefill}{black}    %
\newcommand{\qbadgeopacity}{0.40}  %
\newcommand{\qbadgetextcolor}{white}
\newcommand{\qbadgexsep}{1.2pt}    %
\newcommand{\qbadgeysep}{0.8pt}    %
\newcommand{\qbadgeradius}{1pt}    %
\newcommand{\qbadgefontsize}{10pt} %
\tikzset{tilebadge/.style={anchor=south east,
    font=\fontsize{\qbadgefontsize}{\qbadgefontsize}\selectfont\qbadgefontstyle,
    text=\qbadgetextcolor, fill=\qbadgefill, fill opacity=\qbadgeopacity, text opacity=1,
    inner xsep=\qbadgexsep, inner ysep=\qbadgeysep, rounded corners=\qbadgeradius}}

\newcommand{\qmethodfont}{\scriptsize}

\newcommand{\qrowH}{\dimexpr2\qtileH+\qpairsep\relax}

\makeatletter
\newlength{\qpoolw}          
\newlength{\qpool@maxw}
\newsavebox{\qpool@box}
\newcommand{\poolfit}[1]{%
  \sbox{\qpool@box}{\adjustbox{max totalheight=\qrowH}{#1}}%
  \ifdim\wd\qpool@box>\qpool@maxw \global\qpool@maxw=\wd\qpool@box\fi
  \ifdim\qpoolw=\z@ \else
    \parbox[c][\qrowH][c]{\qpoolw}{\centering\usebox{\qpool@box}}\fi}
\makeatother
\newcolumntype{P}{>{\centering\arraybackslash}m{\qpoolw}}

\newlength{\qtilecolsep}  
\newlength{\qtilerowsep}  
\newlength{\qpairsep}     
\newlength{\qpoolsep}     
\newlength{\qbandrowsep}  
\newlength{\qtitlerowsep} 
\newcommand{\qtilestretch}{0.667}
\makeatletter
\newcommand{\qsetstretch}{%
  \@tempcnta=\numexpr\dimexpr\qtilerowsep\relax*1000/\dimexpr0.3\baselineskip\relax\relax
  \edef\qtilestretch{\strip@pt\dimexpr\@tempcnta sp*65536/1000\relax}}
\makeatother
\newlength{\bandpadxlen}  

\newlength{\subtabvsep}  

\newlength{\qtileH}
\newlength{\qtilew}
\newlength{\qblockw}

\newcommand{\qnafont}{\qmethodfont\color{black!45}}
\newcommand{\natile}[1][]{%
  \def\qna@h{\qtileH}%
  \if\relax\detokenize{#1}\relax\else\sbox0{\borderedimg{#1}}\def\qna@h{\dimexpr\ht0+\dp0\relax}\fi
  \parbox[b][\qna@h][c]{\linewidth}{\centering\qnafont N.A.}}

\newlength{\qlabelw}     
\newlength{\qlabelgap}   
\newlength{\qlabelgapin} 
\newcolumntype{S}{>{\centering\arraybackslash}m{\qlabelw}@{\hspace{\qlabelgapin}}}
\newcolumntype{R}{>{\centering\arraybackslash}m{\qlabelw}@{\hspace{\qlabelgapin}}}
\makeatletter
\newsavebox{\qgrid@box}
\newcommand{\qgridbox}[1]{%
  \setlength{\qlabelgapin}{\qlabelgap}%
  \setlength{\qpoolw}{0pt}\global\qpool@maxw=\z@
  \sbox{\qgrid@box}{#1}%
  \ifdim\qpool@maxw>\z@ \setlength{\qpoolw}{\qpool@maxw}\sbox{\qgrid@box}{#1}\fi
  \ifdim\wd\qgrid@box>\linewidth
    \setlength{\qlabelgapin}{\dimexpr\qlabelgap*\numexpr\dimexpr\wd\qgrid@box\relax*1000/\dimexpr\linewidth\relax\relax/1000\relax}%
    \sbox{\qgrid@box}{#1}%
  \fi
  \usebox{\qgrid@box}}
\makeatother

\newcommand{\qtileradius}{4pt}       %
\newcommand{\qtileborderwidth}{0.4pt}
\makeatletter
\newcommand{\qtile@src}{}
\newcommand{\qtile@try}[1]{\IfFileExists{#1}{\def\qtile@src{#1}}{}}
\newcommand{\qtile@resolve}[1]{%
  \def\qtile@src{figures/placeholder-square}%
  \qtile@try{figures/#1.png}\qtile@try{figures/#1.jpg}%
  \qtile@try{#1.png}\qtile@try{#1.jpg}%
  \qtile@try{figures/#1}\qtile@try{#1}%
}
\newcommand{\borderedimg}[2][\linewidth]{%
  \qtile@resolve{#2}%
  \begin{tikzpicture}[inner sep=0pt, outer sep=0pt]%
    \node[draw=black!25, line width=\qtileborderwidth, rounded corners=\qtileradius, fill=white,
          path picture={\node at (path picture bounding box.center)
            {\includegraphics[width=\dimexpr#1-0.8pt\relax]{\qtile@src}};}]
         {\phantom{\includegraphics[width=\dimexpr#1-0.8pt\relax]{\qtile@src}}};%
  \end{tikzpicture}%
}
\makeatother

\newcommand{\bandpadx}{\bandpadxlen}  %
\newcommand{\bandpadtop}{2pt}  %
\newcommand{\bandpadbot}{0pt}  %
\newcommand{\bandradius}{4pt}  %
\makeatletter
\newcommand{\bandmark}[1]{\pgfsys@markposition{#1}}

\newcommand{\colband}[3]{%
  \begin{tikzpicture}[remember picture, overlay]
    \ifcsname pgf@sys@pdf@mark@pos@#2-#3-l\endcsname
    \ifcsname pgf@sys@pdf@mark@pos@#2-t\endcsname
    \ifcsname pgf@sys@pdf@mark@pos@#2-b\endcsname
      \pgfsys@getposition{#2-#3-l}{\band@l}\pgfsys@getposition{#2-#3-r}{\band@r}%
      \pgfsys@getposition{#2-t}{\band@t}\pgfsys@getposition{#2-b}{\band@b}%
      \pgfsys@getposition{\pgfpictureid}{\band@o}%
      \pgf@process{\pgfpointdiff{\band@o}{\band@l}}\edef\band@lx{\the\pgf@x}%
      \pgf@process{\pgfpointdiff{\band@o}{\band@r}}\edef\band@rx{\the\pgf@x}%
      \pgf@process{\pgfpointdiff{\band@o}{\band@t}}\edef\band@ty{\the\pgf@y}%
      \pgf@process{\pgfpointdiff{\band@o}{\band@b}}\edef\band@by{\the\pgf@y}%
      \fill[#1, rounded corners=\bandradius]
        ([xshift=-\bandpadx, yshift=\bandpadtop] \band@lx, \band@ty) rectangle
        ([xshift=\bandpadx, yshift=-\bandpadbot] \band@rx, \band@by);
    \fi\fi\fi
  \end{tikzpicture}%
}
\makeatother

\definecolor{basecolor}{RGB}{250,196,60}

\newcommand{\best}[1]{\textbf{#1}}
\newcommand{\secondbest}[1]{\underline{#1}}

\definecolor{first}{rgb}{1, 0.7, 0.7}    %
\definecolor{second}{rgb}{1, 0.85, 0.7}  %
\definecolor{third}{rgb}{1, 1, 0.7}      %
\definecolor{highlight}{rgb}{0.8, 1, 0.8} %

\definecolor{customred}{rgb}{0.72, 0.33, 0.31}
\definecolor{customgreen}{rgb}{0.51, 0.70, 0.4}
\definecolor{customblue}{rgb}{0.423, 0.5568, 0.7490}

\definecolor{oursaccent}{rgb}{0.423, 0.5568, 0.7490}  %
\definecolor{oursaccentnew}{rgb}{0.72, 0.33, 0.31}  %

\definecolor{greenwater}{rgb}{0.23, 0.72, 0.67}
\definecolor{salmon}{rgb}{0.89, 0.5, 0.38}

\providecommand{\cmark}{\textcolor{green!55!black}{\ensuremath{\checkmark}}}
\providecommand{\xmark}{\textcolor{black!40}{\ensuremath{\times}}}
\providecommand{\warnmark}{\textsuperscript{\ensuremath{\triangle}}}

\newcommand{\bakemargin}{6pt}
\makeatletter
\newcommand{\bakedpage}[3]{%
  \@namedef{bakedpage@#1}{#2}\@namedef{bakedcites@#1}{#3}}
\newcommand{\bakeddepth}[2]{\@namedef{bakeddepth@#1}{#2}}
\bakedpage{pipeline}{1}{}%
\bakedpage{visibility_dda}{2}{}%
\bakedpage{voxel_correspondence}{3}{}%
\bakedpage{qualitative_co3d_appearance}{4}{}%
\bakedpage{qualitative_co3d}{5}{chang2025reconviagen,huang2025cupid,kerbl3Dgaussians,li2026mv,li2026pixal3d,sam3dteam2025sam3d3dfyimages,xiang2025trellis2,yan2026generative3dgaussianslearned,zadaianchuk2026recgen,zhou2026stream3d}%
\bakedpage{qualitative_completion}{6}{}%
\bakedpage{qualitative_davis_appearance}{7}{}%
\bakedpage{qualitative_davis}{8}{karhade2025any4d,liang2025himor,litman2026lift4d,sam3dteam2025sam3d3dfyimages}%
\bakedpage{qualitative_failures}{9}{karhade2025any4d,liang2025himor,litman2026lift4d,sam3dteam2025sam3d3dfyimages}%
\bakedpage{qualitative_gso_appearance_main}{10}{}%
\bakedpage{qualitative_gso_appearance}{11}{}%
\bakedpage{qualitative_gso}{12}{chang2025reconviagen,huang2025cupid,kerbl3Dgaussians,li2026mv,li2026pixal3d,sam3dteam2025sam3d3dfyimages,xiang2025trellis2,yan2026generative3dgaussianslearned,zadaianchuk2026recgen,zhou2026stream3d}%
\bakedpage{qualitative_oab_appearance}{13}{}%
\bakedpage{qualitative_oab}{14}{liang2025himor,litman2026lift4d,sam3dteam2025sam3d3dfyimages}%
\bakedpage{qualitative_mixed_static}{15}{chang2025reconviagen,huang2025cupid,kerbl3Dgaussians,li2026mv,li2026pixal3d,sam3dteam2025sam3d3dfyimages,xiang2025trellis2,yan2026generative3dgaussianslearned,zadaianchuk2026recgen}%
\bakedpage{qualitative_mixed_dynamic}{16}{karhade2025any4d,liang2025himor,litman2026lift4d,sam3dteam2025sam3d3dfyimages}%
\bakedpage{benchmarks_static}{17}{}%
\bakedpage{benchmarks_dynamic}{18}{}%
\bakedpage{ablations_static}{19}{}%
\bakedpage{ablations_dynamic}{20}{}%
\bakedpage{ablation_poses_full}{21}{}%
\bakedpage{teaser}{22}{}%
\bakeddepth{pipeline}{85.21692pt}%
\bakeddepth{visibility_dda}{0.0pt}%
\bakeddepth{voxel_correspondence}{0.0pt}%
\bakeddepth{qualitative_co3d_appearance}{198.05441pt}%
\bakeddepth{qualitative_co3d}{102.0108pt}%
\bakeddepth{qualitative_completion}{24.13489pt}%
\bakeddepth{qualitative_davis_appearance}{152.2626pt}%
\bakeddepth{qualitative_davis}{114.10616pt}%
\bakeddepth{qualitative_failures}{71.20215pt}%
\bakeddepth{qualitative_gso_appearance_main}{45.99736pt}%
\bakeddepth{qualitative_gso_appearance}{243.14236pt}%
\bakeddepth{qualitative_gso}{125.99622pt}%
\bakeddepth{qualitative_oab_appearance}{221.09853pt}%
\bakeddepth{qualitative_oab}{104.73213pt}%
\bakeddepth{qualitative_mixed_static}{71.59068pt}%
\bakeddepth{qualitative_mixed_dynamic}{71.20215pt}%
\bakeddepth{benchmarks_static}{77.64912pt}%
\bakeddepth{benchmarks_dynamic}{35.47316pt}%
\bakeddepth{ablations_static}{52.33755pt}%
\bakeddepth{ablations_dynamic}{39.19809pt}%
\bakeddepth{ablation_poses_full}{40.56912pt}%
\bakeddepth{teaser}{87.39719pt}%

\newcommand{\bakedfig}[1]{%
  \@ifundefined{bakedpage@#1}{\PackageError{bake}{#1 not baked; run bake/bake.sh}{}}{}%
  \edef\bake@keys{\@nameuse{bakedcites@#1}}%
  \ifx\bake@keys\@empty\else
    \expandafter\nocite\expandafter{\bake@keys}%
    \ifdefined\Hy@backout\@for\bake@k:=\bake@keys\do{\Hy@backout{\bake@k}}\fi%
  \fi
  \begingroup\edef\x{\endgroup\noexpand\raisebox{-\@nameuse{bakeddepth@#1}}{%
    \noexpand\includegraphics[page=\@nameuse{bakedpage@#1},
    trim=\bakemargin\space\bakemargin\space\bakemargin\space\bakemargin]{baked/figures.pdf}}}\x}
\makeatother

\newif\ifteasercallouts
\teasercalloutsfalse

\newsavebox{\cbbox}
\newlength{\cbtriml}
\newcommand{\coverbox}[3]{%
  \sbox\cbbox{\includegraphics[width=#1]{#3}}%
  \ifdim\ht\cbbox<#2\relax
    \sbox\cbbox{\includegraphics[height=#2]{#3}}%
    \setlength{\cbtriml}{\dimexpr(\wd\cbbox-#1)/2\relax}%
    \edef\cbclip{\the\cbtriml\space 0pt \the\cbtriml\space 0pt}%
    \expandafter\clipbox\expandafter{\cbclip}{\includegraphics[height=#2]{#3}}%
  \else
    \setlength{\cbtriml}{\dimexpr(\ht\cbbox-#2)/2\relax}%
    \edef\cbclip{0pt \the\cbtriml\space 0pt \the\cbtriml}%
    \expandafter\clipbox\expandafter{\cbclip}{\includegraphics[width=#1]{#3}}%
  \fi
}

\definecolor{srbgreen}{rgb}{0.0,0.5,0.0}

\usepackage{soul}

\newcommand{\scene}[1]{\texttt{\detokenize{#1}}}

\makeatletter
\renewcommand\paragraph{\@startsection{paragraph}{4}{\z@}%
  {1.2ex \@plus .3ex \@minus .2ex}{-0.7em}{\normalfont\normalsize\bfseries}}
\makeatother
\setlist{topsep=2pt, itemsep=1pt, parsep=0pt}

\newcommand{\SO}{\mathrm{SO}}
\newcommand{\Normal}{\mathcal{N}}

\newcommand{\view}{j}                    %
\newcommand{\numviews}{n_{\mathrm{v}}}   %
\newcommand{\numframes}{n_{\mathrm{f}}}  %
\newcommand{\numobs}{n_{\mathrm{o}}}     %
\newcommand{\eye}{\mathbf{I}}            %

\newcommand{\verts}{V}

\newcommand{\defm}{\boldsymbol{\delta}}   %
 
\newcommand{\layout}{\rho}
\newcommand{\rot}{R}
\newcommand{\trans}{\bt}
\newcommand{\scale}{s}

\newcommand{\decGauss}{\mathcal{D}_{g}}
\newcommand{\decLayout}{\mathcal{D}_{\rho}}
\newcommand{\decOcc}{\mathcal{D}_{o}}          %

\newcommand{\lat}{\bx}                  %

\newcommand{\slat}{Z}                   %
\newcommand{\voxels}{O}
\newcommand{\gauss}{\mathcal{G}}        %
\newcommand{\loss}{\mathcal{L}}         %
\newcommand{\render}{\mathcal{R}}       %
\newcommand{\vis}{\nu}                  %
\newcommand{\patchset}{\Pi}             %

\newcommand{\ftime}{\tau}
\newcommand{\fstep}{\Delta\ftime}
\newcommand{\vel}{\bv}
\newcommand{\velnet}{\bv_\theta}
\newcommand{\cond}{\bc}

\newcommand{\img}{I}                    %
\newcommand{\mask}{M}                    %
\newcommand{\pmap}{P}                    %
\newcommand{\extr}{E}                    %
\newcommand{\intr}{K}                    %
\newcommand{\depth}{D}                   %
\newcommand{\mediandepth}{d}             %

\newcommand{\lossrgb}{\loss_{\mathrm{rgb}}}     %
\newcommand{\lossssim}{\loss_{\mathrm{ssim}}}   %
\newcommand{\losslpips}{\loss_{\mathrm{lpips}}} %
\newcommand{\lossdepth}{\loss_{\mathrm{depth}}} %
\newcommand{\lossrender}{\loss_{\mathrm{rend}}} %
\newcommand{\lossttt}{\loss_{\mathrm{ttr}}} %
\newcommand{\anchorw}{\mu}                   %
\newcommand{\lw}{\omega}                     %

\newcommand{\fusetemp}{\gamma}      %

\newcommand{\method}{{GenIA}}

\begin{document}

\makeatletter
\g@addto@macro\@maketitle{\vspace{6pt}\begingroup
\setlength{\lineskip}{0pt}%
\centering
\bakedfig{teaser}
\par%
\captionsetup{hypcap=false}%
\captionof{figure}{
\textbf{\method{} reconstructs complete, detailed, and input-aligned 3D objects
from monocular images, multi-view images, and monocular videos.}
Our inference-time framework aligns the frozen SAM3D~\cite{sam3dteam2025sam3d3dfyimages} model to the inputs using geometric
and photometric constraints, improving pose and appearance consistency while preserving
generative completion. 
These contributions extend SAM3D beyond the single-view static setting it was trained for, enabling multi-view reconstruction.
Given a temporal coarse shape as input, they enable dynamic reconstruction with improved image quality and stable tracking.
We show object reconstructions from custom and DAVIS~\cite{Perazzi_CVPR_2016} scenes, alongside representative baselines (SAM3D~\cite{sam3dteam2025sam3d3dfyimages}, CUPID~\cite{huang2025cupid}, MV-SAM3D~\cite{li2026mv}, RecGen~\cite{zadaianchuk2026recgen}, Lift4D~\cite{litman2026lift4d}) and input images.
}
\label{fig:teaser}%
\vspace{10pt}
\endgroup
}
\makeatother
\maketitle

\begin{abstract}
Reconstructing complete 3D object assets from monocular or sparse multi-view observations remains challenging.
Generative 3D foundation models can complete object geometry beyond the observed views,
but their predictions may not faithfully reproduce the observed geometry, appearance, or
pose. 
We introduce \method{}, a framework for
test-time input-aligned generation that grounds SAM3D's generative prior in
geometric and photometric observations without retraining the foundation model. 
We improve object pose by deriving translation and scale from geometry while retaining the learned rotation prior, and align appearance through visibility-biased attention, cross-observation fusion, and differentiable rendering guidance during denoising.
An optional post-denoising refinement further adapts the appearance latent, lightweight decoder
adapters, and object placement to the observations.
Our framework also supports externally supplied geometry; when given temporal shapes of dynamic objects, it recovers a shared, input-aligned canonical appearance and stable world-space placement.
Across synthetic and real benchmarks, \method{} improves pose prediction and object reconstruction from monocular, multi-view, and dynamic inputs, outperforming recent optimization-based, per-frame image-to-3D, and video-to-4D methods.
Our project page is available at \texttt{\url{https://facebookresearch.github.io/GenIA}}.
\end{abstract}

\section{Introduction}
\label{sec:introduction}
Photorealistic 3D reconstruction requires recovering complete geometry, faithful appearance, and correct placement from one or more observations.
For dynamic objects, these properties must also remain coherent over time. 
Such assets are central to content creation, immersive media, robotics simulation, and embodied AI, yet remain difficult to recover from casual images or videos. 
The core challenge is to combine generative priors that complete unseen regions with reconstruction constraints that faithfully reproduce the observed evidence.

Recent 3D foundation models learn strong generative priors from large-scale data, but often trade fidelity to the observed inputs for generative completeness.
SAM3D~\cite{sam3dteam2025sam3d3dfyimages}, for example, predicts complete object geometry, appearance, and rigid pose from a single image, but its geometry and appearance may not tightly reproduce the observations, and its pose estimates can be inaccurate. 
At the other extreme, pixel-aligned predictors like Pixal3D~\cite{li2026pixal3d} faithfully reconstruct observed geometry and appearance but offer weaker completion of unseen regions and typically do not recover explicit object pose. 
Another line of work narrows this gap by retraining generative models with reconstruction cues: ReconViaGen~\cite{chang2025reconviagen} conditions structured-latent generation on multi-view features and steers sampling toward the inputs, while CUPID~\cite{huang2025cupid} jointly recovers camera pose and injects pixel-aligned features into a single-image generator. 
However, both specialize the generator to a particular input setting at training time, and neither addresses dynamic objects.

In this work, we instead align the generative prior to its inputs at inference time, without retraining the foundation model.
We call this approach test-time \textbf{I}nput-\textbf{A}ligned \textbf{Gen}eration (\method{}).
Geometric and photometric signals constrain the observed regions, while the generative is steered toward consistent completion of what remains unseen.
The same mechanism spans monocular, multi-view, and dynamic inputs.

\method{} aligns generation to the observations at three levels.
First, we recover translation and scale using depth-derived cues while retaining the generative rotation prior.
Second, we condition appearance on geometry and visibility, directing local features toward the image regions that observe them and fusing predictions across views or frames with visibility weighting.
Third, differentiable rendering provides both in-trajectory guidance and optional
post-denoising refinement.
Our grounding mechanisms are geometry-predictor agnostic and therefore also support externally supplied shapes. 
We demonstrate this with ActionMesh~\cite{ActionMesh2026}, which provides canonical dynamic geometry but neither world-space placement nor appearance; \method{} complements it with input-aligned canonical appearance and temporally stable rigid placement.

\method{} improves pose and object reconstruction from monocular and multi-view inputs without retraining the underlying foundation model, and produces temporally consistent dynamic video reconstructions. It outperforms per-frame image-to-3D, optimization-based, and recent video-to-4D methods on both real and synthetic benchmarks.

Our contributions are:
\begin{enumerate}
    \item An inference-time framework that aligns SAM3D's generative prior with geometric and photometric observations without retraining the foundation model, optionally supporting externally supplied temporal geometry.
    \item A visibility-aware appearance approach combining attention bias, rendering guidance, and cross-observation fusion for monocular, multi-view, and dynamic inputs, all applied to the frozen generator.
    \item A geometry-grounded object placement strategy with sequence-level rotation
    aggregation for improved and temporally stable poses.
    \item An optional post-denoising refinement that jointly optimizes the appearance latent, lightweight decoder adapters, and object placement via differentiable rendering, bridging the gap between the frozen decoder's capacity and the observed appearance while preserving the pretrained prior.
\end{enumerate}

\section{Related work}
\label{sec:related_works}

\subsection{3D reconstruction and generation}

High-fidelity 3D reconstruction was long dominated by \emph{per-scene optimization}, from neural radiance fields~\cite{mildenhall2020nerf,mueller2022instant,wang2021neus,yariv2021volsdf} to 3D Gaussian Splatting~\cite{kerbl3Dgaussians}. Monocular depth estimation has simultaneously progressed from specialized predictors~\cite{ranftl2020midas} to foundation-scale models~\cite{yang2024depthanything,ke2025marigold,bochkovskii2024depthpro}, while general \emph{3D foundation models} directly regress dense geometry and cameras from uncalibrated image collections~\cite{dust3r_cvpr24,mast3r_eccv24,wang2025vggt,wang2025pi,keetha2026mapanything}.
Building on these advances, \emph{feed-forward 3DGS} models recover scene-level Gaussian representations from calibrated views~\cite{charatan23pixelsplat,chen2024mvsplat,xu2024depthsplat} or uncalibrated image collections~\cite{liu2025worldmirror,depthanything3,jiang2025anysplat,gao2026pixworld,Lars2025Sharp}, while recurrent refinement methods improve these predictions through rendering feedback~\cite{xu2025resplat,pearl2026learn2splat}.
While faithful to the observed imagery, these methods reconstruct only visible geometry and do not recover complete, object-centric assets with explicit world-space grounding.

Complementary to reconstruction, \emph{object-centric generation} infers complete geometry and appearance conditioned on text or sparse observations. 
Early approaches distill pretrained 2D diffusion priors through score distillation~\cite{poole2022dreamfusion,lin2023magic3d,tang2024dreamgaussian} or improve the multi-view consistency of image-conditioned generation~\cite{liu2023zero,liu2024syncdreamer,long2024wonder3d}, while large reconstruction models~\cite{hong2024lrm,xu2024instantmesh,LaRa2024ECCV} bridge the two paradigms by regressing complete objects from single or sparse views.
Recent methods instead operate directly in native 3D representations, learning structured latent spaces over voxels, meshes, or Gaussian primitives~\cite{xiang2024structured,xiang2025trellis2,li2025triposg,zhao2025hunyuan3d,yan2026generative3dgaussianslearned}. 
Pixal3D~\cite{li2026pixal3d} departs from canonical-space generation by predicting pixel-aligned assets through explicit image-to-geometry correspondence.
ReconViaGen~\cite{chang2025reconviagen} and CUPID~\cite{huang2025cupid} narrow the gap
between generation and reconstruction through stronger learned alignment to the input:
ReconViaGen targets multi-view reconstruction, while CUPID jointly models object
generation and camera pose from a single image. These approaches rely on specialized
architectures and incorporate alignment directly into the training process; in contrast, we ground pose and appearance at inference
time on a frozen generator, with the same framework spanning monocular, multi-view,
and dynamic inputs.

Our work builds on SAM3D~\cite{sam3dteam2025sam3d3dfyimages}, which predicts complete
object geometry, appearance, and pose from a single image using large-scale
generative priors. Subsequent work accelerates inference~\cite{feng2026fastsam3d3dfyimagesfaster}, extends generation to multiple
views~\cite{li2026mv,zhou2026stream3d}, or retrains the pipeline for pose-conditioned appearance~\cite{zadaianchuk2026recgen}.
Nevertheless, these approaches remain driven primarily by learned priors and do not explicitly enforce consistency with observed images during inference.

Our work is complementary to this line of research. Rather than improving the underlying generative prior, we ground SAM3D's pose and appearance in observations through depth-based initialization, visibility-aware reasoning, differentiable rendering guidance, and optional test-time refinement. 

\subsection{4D reconstruction and generation}

\emph{Dynamic reconstruction} extends static reconstruction by enforcing temporal consistency across geometry, appearance, and motion. Early deformable neural rendering methods~\cite{pumarola2020dnerf,park2021nerfies,park2021hypernerf,TiNeuVox,luiten2023dynamic,yang20244dgs,Wu_2024_CVPR,yang2023deformable3dgs,som2024,lei2025mosca} optimize dynamic scenes for each video, introducing explicit time-varying Gaussian representations through persistent tracking, deformation fields, or compact motion models. 
Per-scene monocular methods improve dynamic reconstruction via temporally consistent meshes~\cite{liu2024dgmesh}, better motion representations~\cite{liang2025himor, wu2025orientation}, or diffusion-based inpainting of novel views~\cite{chen2025cognvs}.
Other methods optimize over learned priors~\cite{zhang2025monst3r, chen2025easi3r, li2025megasam, Mazur_2026_CVPR} or predict 4D geometry and motion directly~\cite{zhang2026d4rt, karhade2025any4d, mi2025one4d}.
However, optimization-based methods are costly per sequence, and all of these methods recover geometry only for surfaces observed in the input, leaving unseen regions incomplete.

\emph{Generative video-to-4D} introduces learned priors to aid or replace per-sequence optimization. Early approaches distill pretrained 2D diffusion models into dynamic representations and supervise 4D optimization using generated multi-view videos~\cite{jiang2024consistent4d,li2024dreammesh4d,xie2024sv4d}. 
Feed-forward methods directly predict temporally complete geometry from monocular videos~\cite{ren2024l4gm,malik2025velox}, while more recent work unifies reconstruction and generation in learned spatio-temporal latent spaces~\cite{zhang2025gaussian,yenphraphai2025shapegen4d,ActionMesh2026,samuel2026fast4dmesh}.
Other works combine image-to-3D priors with per-sequence deformation optimization~\cite{chen2025v2m4,litman2026lift4d,tumanyan2026drops, sang2026twosquared}.

Rather than generating 4D geometry itself, we leverage temporally coherent geometry from existing backbones such as ActionMesh and recover the complementary components they lack: observation-grounded appearance and stable world-space trajectories, without dataset-level retraining.

\begin{figure*}[!t]
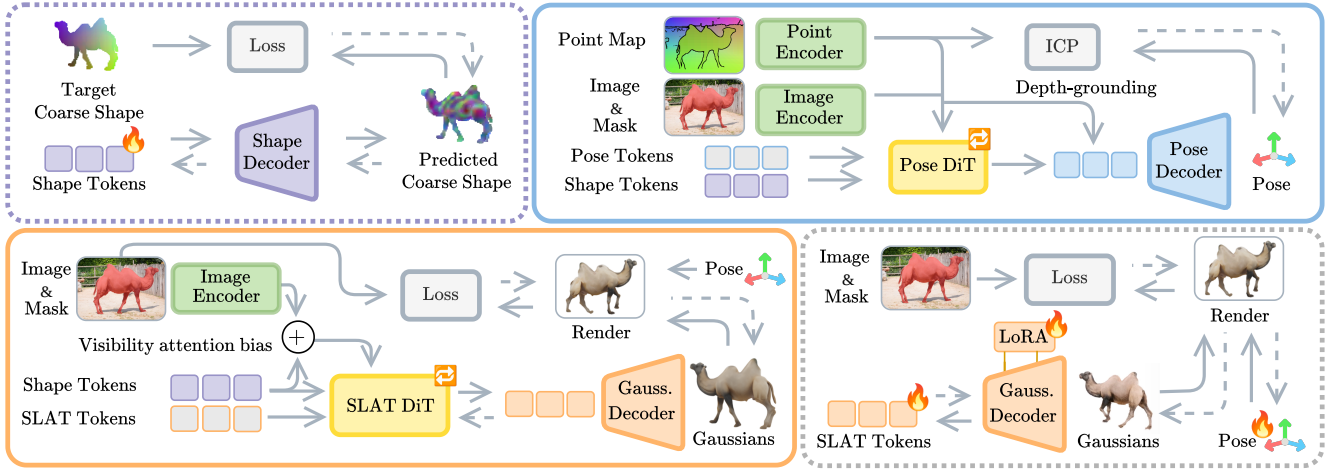
%
\centering
\bakedfig{pipeline}

\caption{\textbf{Our pipeline.} 
Left to right, top to bottom, we
obtain the coarse shape from SAM3D's first pass or (\emph{optionally}) inject an external one into SAM3D's shape tokens (\cref{sec:method:shape_injection}),
denoise each object's per-frame pose over that fixed shape (\cref{sec:method:poses_init}),
predict observation-aligned appearance as SLAT features (\cref{sec:method:appearance}), and
(\emph{optionally}) do test-time refinement of the reconstruction against the input views (\cref{sec:method:ttt}).
Dotted arrows denote gradient flow. We illustrate the single-view setting; multi-view
and dynamic extensions are described in the text.
}
\label{fig:pipeline}%
\end{figure*}

\section{Preliminaries}
\label{sec:preliminaries}

Each observation provides an image $\img$, an object mask $\mask$, a depth map $\depth$, the camera intrinsics $\intr$ and extrinsics $\extr$.

\subsection{SAM3D}
\label{sec:prelim:sam3d}
We build on SAM3D~\cite{sam3dteam2025sam3d3dfyimages}, which operates on a single observation.
SAM3D predicts a coarse occupancy grid $O$, appearance $A$, and pose
$\layout\coloneqq(\rot,\trans,\scale)$, mapping local object coordinates to the camera
frame. 
These are predicted conditioned on a single observation
$(\img,\mask,\intr,\depth)$.
$\intr$ and $\depth$ are internally predicted by a monocular depth estimator (MoGe~\cite{wang2026moge}) and converted into a conditioning pointmap $\pmap$. The camera extrinsics $\extr$ are assumed to be identity, so world and camera space coincide.

The model is a two-stage rectified flow-matching pipeline that denoises a set of modality $m$ latents $\lat^m$, conditioned on input features $\cond$.
The first stage predicts the coarse shape $O$ and the pose $\layout$; the second predicts a structured latent (SLAT) $\slat$ that jointly encodes fine geometry $S$ and appearance $A$.
Both stages use 25-step rectified conditional flow matching with a linear interpolant
$\lat^{m}_{\ftime} \coloneqq \ftime\, \lat^{m}_{1} + (1 - \ftime)\, \lat^{m}_{0}$, $\ftime \in [0,1]$, 
$\lat^{m}_{0} \sim \Normal(\mathbf{0}, \eye)$, target velocity
$\vel^m\coloneqq\dot{\lat}^{m}_{\ftime} = \lat^{m}_{1} - \lat^{m}_{0}$, and learned velocity
$\velnet^{m}(\lat^{m}_{\ftime}, \cond, \ftime)$.
Generation then integrates the ODE
\begin{equation}
  \lat_{\ftime + \fstep}^m
  \;=\;
  \lat_{\ftime}^m \;+\; \velnet^m(\lat_{\ftime}^m, \cond, \ftime)\, \fstep\,.
  \label{eq:flow-matching-ode}
\end{equation}
The velocity is classifier-free guided: each step evaluates a conditioned pass and a condition-free pass, in which the conditioning $\cond$ is replaced by a null embedding $\varnothing$, and extrapolates between them with guidance strength $w$:
\begin{equation}
  \velnet^m \;\coloneqq\; (1 + w)\, \velnet^m(\lat_{\ftime}^m, \cond, \ftime) \;-\; w\, \velnet^m(\lat_{\ftime}^m, \varnothing, \ftime)
  \label{eq:cfg}
\end{equation}

\paragraph{Coarse geometry stage.}
DINOv2~\cite{oquab2024dinov2learningrobustvisual} encodes the image--mask pair at full and mask-cropped resolution into four token channels, concatenated with separately embedded pointmap tokens into the conditioning $\cond$.
A Mixture-of-Transformers~\cite{liang2025mixtureoftransformers} jointly denoises 
(i)~a coarse, voxelized version of the true object shape as a $16^{3} \times 8$ shape latent $\lat^{\voxels}$, decoded by the structure decoder $\decOcc$ into a $64^{3}$ occupancy grid $\voxels$, 
and (ii)~an object pose as three pose tokens $\lat^{\rot}, \lat^{\trans}, \lat^{\scale}$ decoded by the pose decoder $\decLayout$ into $\layout$.
Information flows unidirectionally from the shape stream to the pose stream through multi-modal self-attention.

\paragraph{Fine geometry and appearance stage.}  
The input images and masks are encoded as in the coarse geometry stage
into the conditioning $\cond$; pointmap tokens are omitted. 
A sparse latent flow transformer denoises the SLAT $\slat$ over the active voxels of the coarse occupancy $\voxels$.
The representation can be decoded with $\decGauss$ to a set of 3D Gaussians $\gauss$ in the object frame defined by the predicted coarse shape $O$.

\section{Method}
\label{sec:method}
Given posed observations, \method{} recovers input-aligned object reconstructions with per-frame world-space poses.
By default, SAM3D jointly denoises coarse shape and rotation. When external geometry is available, we instead inject it into SAM3D's coarse latent representation (\cref{sec:method:shape_injection}) and keep the shape fixed while denoising rotation.
We then derive translation and scale from depth while retaining SAM3D's rotation prior (\cref{sec:method:poses_init}), recover a shared appearance representation using visibility and rendering constraints (\cref{sec:method:appearance}), and optionally refine the instance representation and world-space pose at test time (\cref{sec:method:ttt}). 
\Cref{fig:pipeline} summarizes the pipeline.
All stages operate at inference time without updating the pretrained SAM3D weights;
optional refinement optimizes only an object-specific appearance latent, a lightweight
low-rank decoder adapter, and pose. 

We index each observation by view $\view\in\{1,\dots,\numviews\}$ and frame
$k\in\{1,\dots,\numframes\}$: $\numviews$ distinguishes monocular ($=1$) from
multi-view ($>1$) inputs, while $\numframes$ distinguishes static ($=1$) from dynamic
($>1$) scenes.
Rather than using MoGe, which provides single-view depth but no camera pose, we use
MapAnything~\cite{keetha2026mapanything} to obtain metric depth and camera calibration
$(\depth_{(k,\view)},\extr_{(k,\view)},\intr_{(k,\view)})$ across all settings.
Our framework remains agnostic to the geometry predictor, supporting any method that provides geometrically consistent pointmaps and camera parameters.

\subsection{\emph{Optional} Shape Injection}
\label{sec:method:shape_injection}

When external geometry $S$ is available, we replace SAM3D's coarse shape prediction by
voxelizing $S$ into an occupancy grid $O$ and inverting it into the clean latent
$\lat^O_1$ expected by SAM3D's frozen occupancy decoder.
The inversion uses a reconstruction objective with moment-matching regularization to
match the empirical distribution of SAM3D shape tokens ($\mathbb{E}[\lat^O_1]\approx 0,
\operatorname{Var}(\lat^O_1)\approx 0.9$).
Pose denoising (\cref{sec:method:poses_init}) is then conditioned on the injected latent; 
without external geometry, injection is skipped and SAM3D denoises coarse shape and pose jointly.

For dynamic inputs, this external geometry additionally specifies an object-local deformation field $\defm$, relating the canonical shape to each frame (details in \cref{sec:supp:shape_injection}).
We invert each
deformed shape while preserving these correspondences.

\subsection{Pose prediction}
\label{sec:method:poses_init}
We use SAM3D's coarse geometry model to predict the reconstructed object's world-space pose $\layout_k\coloneqq(\rot_k,\trans_k,\scale_k)$ at frame $k$.

\paragraph{Denoising.}
Without external geometry, coarse shape and pose are jointly denoised for SAM3D's
default 25 steps. 
With an injected shape latent $\lat^{\voxels}_{k,1}$, the fixed shape conditions the pose integration \eqref{eq:flow-matching-ode} for 10 steps.
For static \emph{multi-view} inputs ($\numviews>1$), we denoise the pose from the first view ($\view=1$).
For \emph{dynamic} objects, the injected shapes already capture non-rigid motion and local rotation, so we only need to recover a single global root orientation $\rot$ rather than one per frame. We find this as the coordinate-wise median of the rotation velocities predicted across frames at each ODE step.

\paragraph{Depth-grounded translation and scale.}
Translation and scale dominate SAM3D's image-space pose errors (ablated in \cref{sec:supp:ablations}). 
We therefore retain the denoised rotation $\rot_k$ and instead estimate translation and scale from geometric cues.
Given the camera-space pointmap $\pmap_k$, object mask $\mask_k$, and intrinsics
$\intr_k=(f_x,f_y,c_x,c_y)$, we measure three quantities: the object's depth $\mediandepth_k$ (the median $z$ of $\pmap_k$ over the mask), its center $(\bar u_k,\bar v_k)$, and its angular size $a_k=\max(\Delta u_k/f_x,\Delta v_k/f_y)$, where $(\Delta u_k,\Delta v_k)$ is the extent of the mask bounding box in pixels.
We then seek the translation and scale at which the posed coarse shape $\voxels$, projected with $\intr_k$, yields matching predictions: a silhouette with angular size $\hat a$, center $(\hat u,\hat v)$, and median visible depth $\hat d$.
We initialize by back-projecting the mask at $\mediandepth_k$,
\begin{equation}
\trans \leftarrow \mediandepth_k
\left(\frac{\bar u_k-c_x}{f_x},\,\frac{\bar v_k-c_y}{f_y},\,1\right),
\qquad
\scale \leftarrow \mediandepth_k\, a_k ,
\label{eq:depth-init}
\end{equation}
We then perform five fixed-point iterations that update, in order, scale, depth, and lateral position, each from the residual of its own predicted quantity:
\begin{equation}
\begin{aligned}
\scale &\leftarrow \scale\,\tfrac{a_k}{\hat a},
\qquad
\trans_z \leftarrow \trans_z + \mediandepth_k - \hat d, \\
(\trans_x,\trans_y) &\leftarrow (\trans_x,\trans_y)
  + \bigl(\tfrac{\bar u_k-\hat u}{f_x},\,\tfrac{\bar v_k-\hat v}{f_y}\bigr)\trans_z,
\end{aligned}
\label{eq:pose-refit}
\end{equation}
The result replaces the denoised translation and scale.
For \emph{multi-view} inputs, translation and scale are estimated per view as above,
mapped to world coordinates with $\extr_{(k,\view)}$, and
the median center and scale define a shared world-space pose. 
For \emph{dynamic} objects, they remain frame-specific.
The object's world-space pose at frame $k$ then follows by composing the
camera-space pose $\layout_{(k,\view)}$ with $\extr_{(k,\view)}$.

\paragraph{Registration.}
Finally, we refine the initialized pose by registering the coarse shape to the metric
input pointmap with a short iterative closest point (ICP) optimization~\cite{besl1992method} (details in \cref{sec:supp:registration}).

\subsection{Appearance prediction}
\label{sec:method:appearance}
With the coarse shape $\voxels$ fixed and the object pose $\layout$ given, the appearance stage predicts a structured latent representation $\slat$ that encodes fine geometry and appearance. 
Conditioned on the voxel grid and the input images, a sparse latent flow transformer denoises $\slat$ through \eqref{eq:flow-matching-ode}; the resulting latent is decoded into a set of 3D Gaussians (\cref{sec:preliminaries}).

Conditioned on one observation, the SAM3D backbone predicts the velocity for the latent trajectory. 
At every denoising step, we first \emph{bias cross-attention} toward image patches that observe each voxel and then apply \emph{rendering guidance} to the resulting velocity prediction.
The per-observation guided velocity predictions are
then fused at every denoising step using \emph{visibility-based weights}, yielding a shared latent
trajectory. 

In the following, we index observations by $b\in\{1,\dots,\numobs\}$: a camera view
in the static multi-view setting ($\numobs{=}\numviews$) and a frame in the dynamic setting
($\numobs{=}\numframes$).

\begin{figure}[t!]
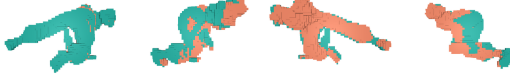

\centering
\bakedfig{visibility_dda}

\caption{%
\textbf{Visibility on the object's coarse shape.}
Left: camera view; orbiting reveals \colorbox{greenwater}{visible} and
\colorbox{salmon}{occluded} voxels.
}
\label{fig:visibility_dda}
\end{figure}

\paragraph{Per-voxel visibility.} 
\label{sec:visibility}
A voxel occluded in a view receives no reliable appearance information from it.
We therefore compute a per-observation voxel visibility mask $\vis_i(b) \in \{0, 1\}$. A voxel $i$ is marked
visible $\vis_i(b)=1$ if the ray from the voxel to the camera reaches the image plane without first intersecting
another occupied cell, accounting for self-occlusion.
\Cref{fig:visibility_dda} visualizes the resulting visibility on a voxelized shape.
For each voxel, we also define $\patchset_i(b)$ as the set of conditioning patches
covering voxel $i$ in observation $b$; if the voxel is not visible, then $\patchset_i(b)=\emptyset$ (details in \cref{sec:supp:visibility-details}).

\paragraph{Visibility attention bias.}
The appearance predictor cross-attends to conditioning patches $\cond$. We bias this attention toward the patches that directly observe the voxels. 
For each 
voxel $i$ and observation $b$, we add $+\alpha$ ($\alpha{=}5$) to the logits of
$\patchset_i(b)$ in the full- and cropped-RGB conditioning groups; occluded voxels
receive no bias. The bias is applied only to the conditioned CFG branch
$\velnet(\lat_{\ftime},\cond,\ftime)$.
Since softmax spans all conditioning tokens, we also subtract a per-voxel offset from
the RGB logits to approximately preserve the attention share of unbiased mask and
global tokens while retaining the $e^\alpha$ visible-patch boost; derivation and exact passive-stream compensation are given in \cref{sec:supp:passive-compensation}.

\begin{figure*}[t]
\centering
\begin{subfigure}{\textwidth}
\centering
\bakedfig{qualitative_mixed_static}

\caption{\footnotesize\textbf{Static datasets} (GSO-30 top, CO3D bottom).}
\label{fig:mixed_static}
\end{subfigure}

\par\vspace{\subtabvsep}
\begin{subfigure}{\textwidth}
\centering
\bakedfig{qualitative_mixed_dynamic}

\caption{\footnotesize\textbf{Dynamic datasets} (ActionBench top, DAVIS bottom).}
\label{fig:mixed_dynamic}
\end{subfigure}

\caption{
\textbf{Qualitative comparison.}
\subref{fig:mixed_static} GSO-30 and CO3D reconstructed from one input view and from
multi-view inputs ($5$ views for GSO-30, $4$ for CO3D).
\subref{fig:mixed_dynamic} ActionBench and DAVIS reconstructed from 16-frame sequences;
for DAVIS, we show the first and last reconstructed frames.
Methods are described in \cref{sec:results:baselines}.
DAVIS provides no ground-truth test views, so its \emph{test} subrow shows synthesized
novel views at the same timestamps.
Each render is badged with per-scene PSNR ($\uparrow$) on \emph{train} views and
CLIP-I ($\uparrow$) on \emph{test} views.
}
\label{fig:mixed_qualitative}
\end{figure*}

\paragraph{Rendering guidance.}
The visibility bias acts through the frozen conditioning pathway; rendering
guidance, instead, directly supervises the decoded reconstruction during the flow trajectory.
Each observation's latent $\slat_b$ is advanced by its own CFG-combined
velocity $\velnet^{(b)}$ \eqref{eq:cfg}; rendering guidance corrects this velocity
independently for every observation.
At each ODE step $\ftime$, we estimate the clean latent by extrapolating along the flow using the flow-matching analogue of Tweedie's formula:
$\hat{\slat}_b(\ftime)=\slat_b(\ftime)+(1-\ftime)\velnet^{(b)}$. This is decoded to
Gaussians and rendered under the fixed pose $\layout_b$ to obtain
$(\hat\img_b,\hat\depth_b)$.
The per-observation rendering loss combines photometric and depth supervision:
\begin{equation}
\begin{aligned}
\lossrender^{(b)} \;=\;{}&
\lossrgb(\hat\img_b,\img_b)
+0.2 \lossssim(\hat\img_b,\img_b) \\
&+ 0.5 \losslpips(\hat\img_b,\img_b)
+0.1 \lossdepth(\hat\depth_b,\depth_b).
\end{aligned}
\label{eq:render-loss-terms}
\end{equation}
Implementation details are in \cref{sec:supp:render-loss}.
The rendering gradient corrects each observation's velocity. To do so, we first rescale the
gradient to the current velocity norm and then apply
$
\velnet^{(b)}
  \leftarrow
  \velnet^{(b)}
  -
  \lambda\,\nabla_{\slat_b(\ftime)}\lossrender^{(b)},
$ where $\lambda$ controls the relative guidance strength.

\paragraph{Visibility-weighted observation fusion.}
For each voxel $i$, we fuse the per-observation latents as
$(\slat)_i=\sum_{b=1}^{\numobs} w_i(b)(\slat_b)_i$, with
\begin{equation}
\resizebox{0.9\linewidth}{!}{$
\tilde w_i(b)
=
\frac{\exp\!\left(\fusetemp\,\vis_i(b)\right)}
{\sum_{b'=1}^{\numobs}\exp\!\left(\fusetemp\,\vis_i(b')\right)},
\qquad
w_i(b)
=
\frac{\max\!\left(\tilde w_i(b),\epsilon\right)}
{\sum_{b'=1}^{\numobs}\max\!\left(\tilde w_i(b'),\epsilon\right)}.
$}
\label{eq:visibility-fusion}
\end{equation}
We use $\fusetemp=30$ and $\epsilon=10^{-3}$. Since visibility is binary, observations
that see a voxel receive equal weight, while the floor retains a small contribution
from occluded observations. Visibility is computed once on fixed coarse geometry
and remains constant throughout denoising.

In the static \emph{multi-view} setting, all observations share the same voxel grid, and
fusion is applied at every ODE step. 

The same formulation extends to \emph{dynamic} objects, using the deformation field
$\defm$ (\cref{sec:method:shape_injection}) to establish correspondences between each
frame-specific grid $\voxels_b$ and the canonical grid $\voxels$. At every ODE step,
per-frame predictions are mapped to the canonical grid and fused with the same
visibility-weighted rule (see \cref{sec:supp:shape_injection}).

\subsection{\emph{Optional} Test-time refinement}
\label{sec:method:ttt}
The frozen generative decoder may only approximately reproduce instance-specific appearance and geometry. 
We therefore optionally refine a low-rank Gaussian-decoder adapter $\Delta\decGauss$, together with the canonical latent
$\slat$ and the world-space
poses $\layout$ for $100$ iterations per object, while keeping all pretrained
parameters fixed (TTR).
Using the rendering objective of \eqref{eq:render-loss-terms}, we optimize these
test-time variables by rendering each observation as
$\render(\decGauss(\slat;\Delta\decGauss),\,\layout_{(k,\view)})$ and minimizing
\begin{equation}
  \lossttt = \sum_{k=1}^{\numframes}\sum_{\view=1}^{\numviews} \lossrender^{(k,\view)} +\; \anchorw\,\lVert \slat - \slat^{0}\rVert^2 ,
  \label{eq:ttt}
\end{equation}
where the final term anchors the optimized latent to its initial prediction
$\slat^0$, preserving the prior's completion of regions the observations do
not constrain.
For dynamic objects, the canonical Gaussians are warped by
$\defm_{c\rightarrow k}$ before rendering each frame.
The latent anchor and decoder adapter are complementary: the anchor preserves the
predicted latent and its completion of unobserved regions, while
$\Delta\decGauss$ provides instance-specific capacity beyond the frozen decoder's
range.
We observe that latent-only refinement, which forces appearance residuals into the globally coupled latent, primarily improves observed views, whereas decoder adaptation also improves held-out views.
Additional details are given in \cref{sec:supp:ttr}.

\begin{table}[t]
  \centering
  \setlength{\tabcolsep}{4pt}

  \begin{subtable}[t]{\linewidth}
    \centering
    \bakedfig{benchmarks_static}
    \caption{\textbf{Static scenes} -- GSO-30 and CO3D}\label{tab:static_full}
  \end{subtable}

  \vspace{\subtabvsep}
  \begin{subtable}[t]{\linewidth}
    \centering
    \bakedfig{benchmarks_dynamic}
    \caption{\textbf{Dynamic scenes} -- ActionBench and DAVIS}\label{tab:dynamic_full}
    \label{tab:davis_table}
  \end{subtable}

  \caption{
    \textbf{Full benchmarks.}
    Per-dataset image quality metrics across methods and,
    where applicable, number of input views (\#V).
    ``--'' denotes metrics not reported by the corresponding method. 
    Baselines, datasets, metrics and geometric backbones used by our method are detailed in \cref{sec:results}.
    TRELLISv2 and Pixal3D predict lighting-disentangled materials and are therefore evaluated qualitatively only.
    \best{Best} and \secondbest{second best} per (\#V, dataset, metric) group are shown in
    bold and underlined, respectively.
    }
  \label{tab:full_benchmarks}
\end{table}

\section{Results}
\label{sec:results}

\paragraph{Baselines.}
\label{sec:results:baselines}
We compare against state-of-the-art methods covering each of our input settings.
\emph{Monocular static:}
We use SAM3D~\cite{sam3dteam2025sam3d3dfyimages} as the geometry predictor and compare
against SAM3D, RecGen~\cite{zadaianchuk2026recgen},
TripoSplat~\cite{yan2026generative3dgaussianslearned}, and CUPID~\cite{huang2025cupid}.
We additionally compare qualitatively with TRELLISv2~\cite{xiang2025trellis2} and
Pixal3D~\cite{li2026pixal3d}, whose lighting-disentangled material predictions lack
scene illumination and therefore do not support comparable photometric re-rendering.
\emph{Multi-view static:}
We use MV-SAM3D~\cite{li2026mv} as the geometry backbone (implemented in our framework and parallelized) and evaluate with varying
numbers of input views. We compare against MV-SAM3D,
STREAM3D~\cite{zhou2026stream3d}, ReconViaGen~\cite{chang2025reconviagen}, and
RecGen~\cite{zadaianchuk2026recgen} (up to two input views).
We further include \emph{Depth-3DGS}~\cite{kerbl3Dgaussians}, a per-scene optimization baseline detailed in \cref{sec:supp:depth3dgs}.
\emph{Monocular dynamic:}
We use ActionMesh~\cite{ActionMesh2026} canonical geometry and deformations, and recover
its missing world-space pose with our pose stage.
We compare against per-frame SAM3D, Lift4D~\cite{litman2026lift4d}, which combines 3D mesh generation with per-scene object optimization, and the per-scene optimization baseline HiMoR~\cite{liang2025himor}.
Additionally, we compare against Any4D~\cite{karhade2025any4d}, which predicts temporal point clouds for 3D tracking.
We run all baselines on the same inputs as our method, including geometric priors where supported.

\paragraph{Datasets.}
\emph{Static:} GSO-30~\cite{kong2024eschernet}, a synthetic multi-view benchmark with
a single object on a white background, and selected CO3D~\cite{reizenstein21co3d} scenes, a real multi-view dataset with one object; both are also evaluated with a single input view.
\emph{Dynamic:} Selected scenes from ActionBench~\cite{ActionMesh2026}, a synthetic monocular benchmark
with a single object on a white background, which we re-render with one held-out view per timestamp;
and DAVIS~\cite{Perazzi_CVPR_2016}, comprising monocular in-the-wild
videos with one object per scene (16 frames at stride 2 over the
first 32 frames).
Exact scenes used and data processing are in \cref{sec:supp:datasets}.

\begin{figure}[!t]
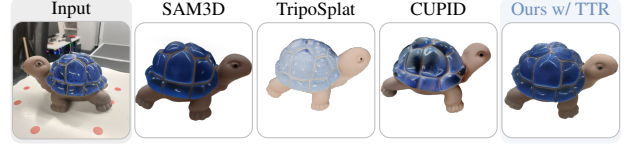

\centering
\bakedfig{qualitative_completion}

\caption{
\textbf{\method{} with TTR completes the unseen surface with colors consistent with the input.}
A real scene rendered from the side opposite the input view. SAM3D completes the object with a smooth, synthetic appearance driven by its generative prior. TripoSplat and CUPID closely match the observed side but degrade on unseen surfaces, with washed-out colors or appearance drift.
}
\label{fig:completion}
\end{figure}

\paragraph{Metrics.}
\label{sec:results:metrics}
We report foreground-masked PSNR on training views and LPIPS/CLIP-I on held-out views. Pixel-aligned metrics are sensitive to pose errors and penalize plausible completions that differ from ground truth, so we additionally report CLIP-I as a coarse semantic measure, despite its limited sensitivity to fine texture and temporal consistency (\cref{sec:supp:metrics}). To reduce pose-induced bias, we refine each method's pose against the inputs while fixing shape and appearance; per-scene optimization methods are exempt (\cref{sec:supp:baseline-poses}). On GSO-30 and ActionBench, co-PSNR further isolates held-out pixels visible in at least one input view.
For DAVIS, which lacks test views, we follow Lift4D~\cite{litman2026lift4d} by synthesizing rotated novel views and averaging CLIP-I against the closest training image.
We evaluate tracking against CoTracker3~\cite{karaev2024cotracker3} pseudo-ground-truth using $\delta_{\mathrm{avg}}$, reprojection error (Err), and AJ; these scores measure agreement with CoTracker3 rather than absolute accuracy (see \cref{sec:supp:pseudo-tracks}).

\begin{figure}[!t]
\centering
\includegraphics[width=\linewidth]{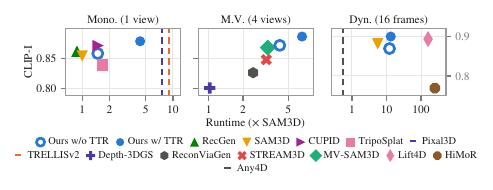}
\caption{%
\textbf{Our method adds a small overhead over SAM3D.}
Median runtime versus mean CLIP-I, with runtime given as a multiple of
SAM3D's, which takes $25$\,s on our hardware (a single A100-40\,GB). 
The two
static panels are scored on CO3D; the dynamic panel is scored on DAVIS.
Without TTR, rendering guidance is the main overhead: runtime is approximately affine in view count, with per-view cost comparable to multi-view baselines. 
TTR adds a largely view-independent cost, while improving CLIP-I. On dynamic sequences, \method{} costs $13\times$ SAM3D's single-image runtime including ActionMesh ($12\times$ without TTR), versus $160\times$ for Lift4D and $250\times$ for HiMoR, and with TTR achieves the highest CLIP-I.
}
\label{fig:time_vs_quality}
\end{figure}

\subsection{Analysis}
\label{sec:results:analysis}
Across datasets, \method{} aligns the frozen SAM3D model to the inputs, improving pose and appearance reconstruction (\cref{tab:full_benchmarks}) while
remaining competitive on runtime (\cref{fig:time_vs_quality}).

In the single-view setting, SAM3D, RecGen, and TRELLISv2 can realistically complete unobserved geometry but often misalign appearance, especially for symmetric objects. Relative to SAM3D, we better resolve these ambiguities and improve color and lighting alignment, with gains extending to held-out views.
Pixal3D, TripoSplat, and CUPID are strongly input-aligned and even achieve higher co-PSNR than our method on synthetic scenes. However, they more frequently produce incomplete or washed-out occluded geometry or textures (see \cref{fig:completion}). On held-out views, our method achieves superior metrics, particularly on real scenes, while qualitatively better preserving backside geometry and appearance.

Similar misalignment persists for MV-SAM3D, STREAM3D, and RecGen in the multi-view setting, with ReconViaGen being the most input-aligned generative baseline. Without TTR, \method{} outperforms all baselines, including ReconViaGen. Depth-3DGS leads training-view PSNR and co-PSNR, but remains incomplete in unobserved regions. TTR further improves fidelity and perceptual quality.

On dynamic scenes, per-frame SAM3D is temporally inconsistent in geometry and appearance and provides no tracking correspondences. 
HiMoR and Any4D reconstruct only observed regions. 
Lift4D, the closest baseline to ours, is noisier and less temporally stable. In comparison, our method inherits the quality of the injected coarse geometry from ActionMesh and results in coherent shape and appearance. 
With TTR, \method{} ranks second to HiMoR on DAVIS training-view PSNR and $\delta_{\mathrm{avg}}$, while achieving the lowest reprojection error and highest CLIP-I. On ActionBench, HiMoR leads training-view PSNR and co-PSNR, as expected for a per-scene optimizer fitted to the input view, while \method{} is on par with per-frame SAM3D and achieves the best held-out LPIPS. Per-frame SAM3D has the highest ActionBench CLIP-I, reflecting its insensitivity to pose alignment.

Our method struggles on ActionBench scenes due to large inter-frame motion, where errors in the coarse geometry or pose can propagate and substantially degrade reconstruction.
Peak GPU memory on CO3D (1 input view) is $19.7$\,GB without TTR versus SAM3D's $18.0$\,GB; TTR increases it to $31.8$\,GB due to back-propagation through the Gaussian decoder.
Additional results are provided in \cref{sec:supp:additional_results}.

\begin{figure}[!t]
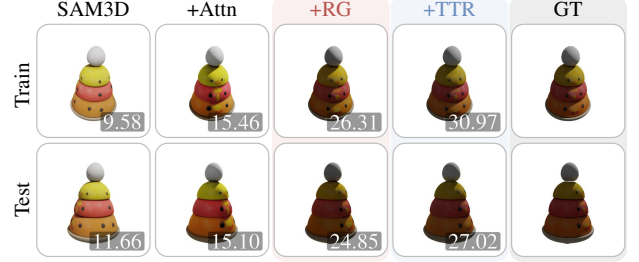

\centering
\bakedfig{qualitative_gso_appearance_main}

\caption{
\textbf{Our appearance prediction better aligns with the input observations.}
Incremental appearance ablation on GSO-30, following \cref{sec:results:ablation}.
Attention bias injects pose information into SAM3D's pose-blind appearance predictor,
helping resolve ambiguous appearance assignment on symmetric coarse geometry.
Rendering guidance further improves color and detail matching, while optional TTR
provides additional gains, including on held-out views.
We show one \emph{train} and one held-out \emph{test} view, badged with PSNR and
co-PSNR ($\uparrow$).
}

\label{fig:gso_appearance_qual_main}
\end{figure}

\subsection{Ablations}
\label{sec:results:ablation}
We ablate our contributions cumulatively;
\Cref{fig:gso_appearance_qual_main} shows the appearance ablation on a static
monocular GSO scene. We use ground-truth coarse geometry to isolate the contribution of
the appearance components. Starting from SAM3D appearance prediction (Base), we add
visibility attention bias (+Attn), rendering guidance (+RG), and test-time refinement
(+TTR), progressively improving both training- and test-view quality.
Full ablations for real and synthetic multi-view and monocular dynamic settings, together with ablations of our pose prediction components, are provided in \cref{sec:supp:ablations}.

\section{Conclusion}
\label{sec:conclusions}
We presented \method{}, an inference-time framework that aligns a frozen generative 3D
prior to its observations. We ground translation and scale in depth while retaining the
prior's rotation, and align appearance through visibility-biased attention, rendering
guidance, and cross-observation fusion.
This improves pose and novel-view quality for static monocular and multi-view inputs
and, with external dynamic geometry, yields temporally coherent reconstructions and
accurate tracks.

\paragraph{Limitations and future work.}
Dynamic geometry currently relies on an external predictor; removing this dependence would enable unified, input-aligned 3D/4D shape prediction. Unlike pose and appearance, coarse shape is not directly grounded in the observed silhouette and depth, as we found the geometry prior difficult to steer reliably.
Rotation remains prior-driven and only weakly corrected by noise-sensitive
registration, while pose errors from masks, depth, and cameras can also degrade the differentiable rendering supervision signal.

\boldparagraph{Acknowledgements}
The authors affiliated with the T\"ubingen AI Center acknowledge the support of the Deutsche Forschungsgemeinschaft (DFG, German Research Foundation) under Germany's Excellence Strategy (EXC number 2064/1, project number 390727645).
Stefano Esposito acknowledges travel support from the European Union's Horizon 2020 research and innovation program under ELISE Grant Agreement No.\ 951847.

{
    \small
    \bibliographystyle{3dv/ieeenat_fullname}
    \bibliography{bib/bibliography_long,bib/bibliography,bib/bibliography_custom}
}

\clearpage
\appendix

\section{Supplementary Material}

\subsection{Extended ablations}
\label{sec:supp:ablations}

\begin{table*}[!t]
  \centering
  \small
  \setlength{\tabcolsep}{3pt}
  \caption{%
    \textbf{Pose ablations across datasets.}
    Four-step cumulative ablation, with each row adding one component of
    \cref{sec:method:poses_init}. We evaluate dynamic scenes (DAVIS and ActionBench with
    ActionMesh geometry) and static GSO-30 single- and multi-view settings with injected
    ground-truth geometry.
    Dynamic settings report reprojection error and relative jitter against the CoTracker3
    pseudo-reference (\cref{sec:supp:pseudo-tracks}), together with masked depth $L_1$.
    GSO-30 additionally measures placement of the injected coarse shape against ground truth
    using Chamfer distance. Jit.\ is relative: $1.0$ matches the reference motion, while
    larger values indicate greater jitter.
    Rotation averaging is inert on static settings by construction, since it averages
    velocities across frames. ICP registration is most beneficial for static multi-view
    alignment, where it halves Chamfer distance; on DAVIS, it improves depth at the cost
    of appearance quality.
  }
  \label{tab:ablation_poses_full}
  \bakedfig{ablation_poses_full}
\end{table*}

\begin{table}[tp]
  \centering
  \setlength{\tabcolsep}{4pt}

  \begin{subtable}[t]{\linewidth}
    \centering
    \bakedfig{ablations_static}
    \caption{\footnotesize\textbf{Static scenes} -- GSO-30 and CO3D}\label{tab:appearance_static}
  \end{subtable}

  \vspace{\subtabvsep}
  \begin{subtable}[t]{\linewidth}
    \centering
    \bakedfig{ablations_dynamic}
    \caption{\footnotesize\textbf{Dynamic scenes} -- ActionBench and DAVIS}\label{tab:appearance_dynamic}
  \end{subtable}

  \caption{
    \textbf{Appearance ablations.}
    Per-dataset novel-view synthesis across the component
    ladder: the columns toggle
    \emph{Attn} (visibility attention bias), \emph{RG} (rendering guidance), and
    \emph{TTR} (test-time refinement); \cmark/\xmark\ mark a component on/off and
    ``--'' a component that does not apply (or a metric not reported).
    Held-out views report co-PSNR where the dataset has co-visibility masks,
    otherwise only LPIPS and CLIP-I.
    All
    \textbf{GSO-30} and \textbf{ActionBench} variants share ground-truth shape and pose.
    \textbf{CO3D} uses SAM3D- and MV-SAM3D-predicted shape and pose.
    \textbf{DAVIS} uses the ActionMesh backbone.
    Metrics are averaged across scenes and views (GSO-30,
    CO3D), views and time (ActionBench), or time (DAVIS). \best{Best} and \secondbest{second best} per (\#V, dataset, metric) group are shown in bold and
    underlined, respectively.
  }
  \label{tab:full_ablations}
\end{table}

\begin{figure*}[t]
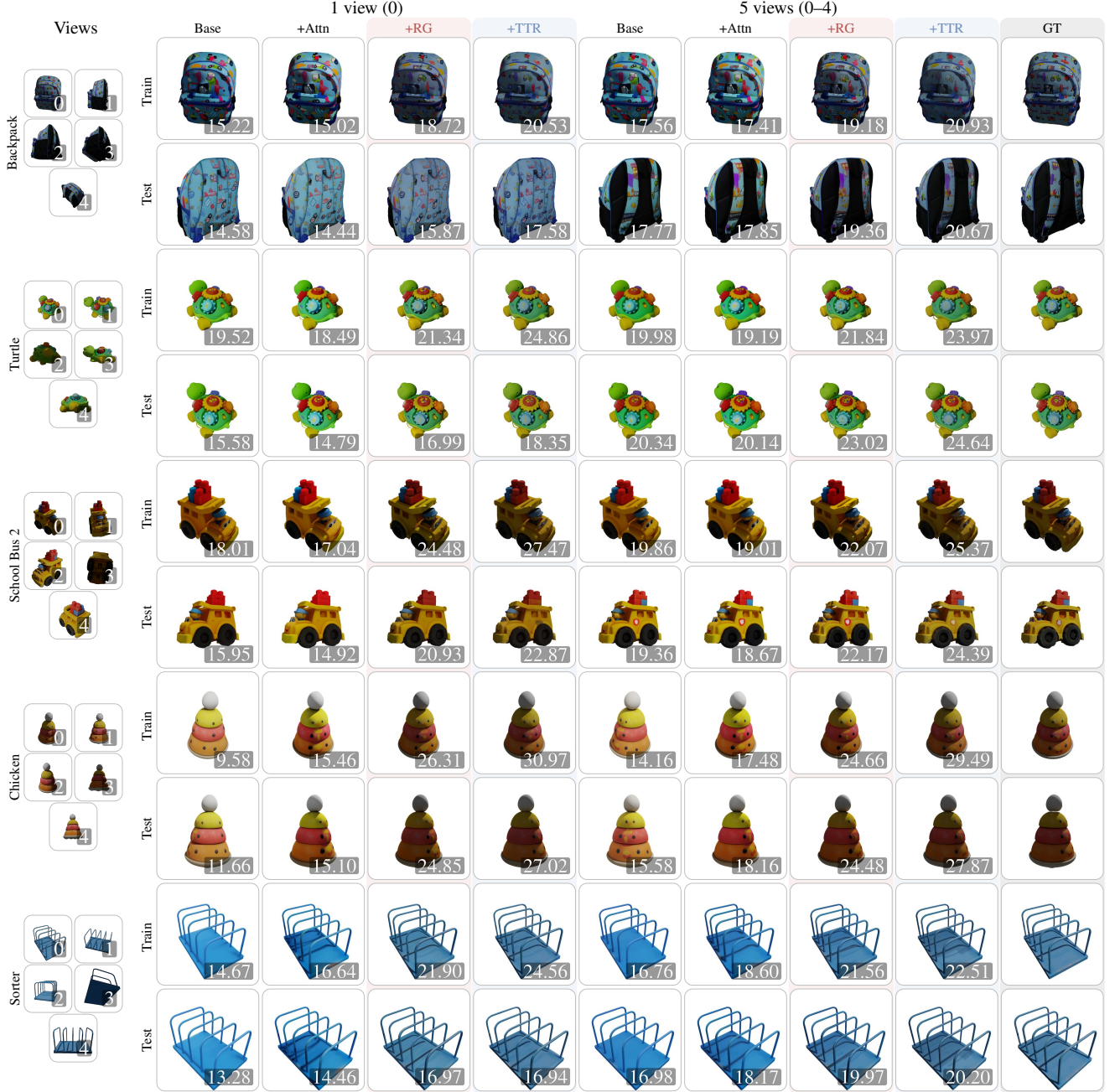

\centering
\bakedfig{qualitative_gso_appearance}

\caption{
\textbf{Qualitative appearance ablation on GSO-30 with ground-truth geometry.}
Every column is our method on the GT-voxelized mesh (shared GT shape and pose), so
geometry is fixed and only \emph{appearance} differs. 
Cumulative ladder, where each column adds one component to the column on its left.
The label names only what that step adds: Base (SAM3D appearance prediction + observation fusion) $\to$
+Attn (visibility attention bias) $\to$ +RG (rendering guidance) $\to$ +TTR
(test-time refinement). 
Each
scene spans two subrows: rendered from the first input (\emph{train}) view and from a
held-out novel (\emph{test}) view, each against its GT. Each render is badged (bottom-right)
with its per-scene appearance score: PSNR ($\uparrow$) on the \emph{train} views and
co-visible PSNR (co-PSNR, $\uparrow$; \cref{sec:results:metrics}) on the
\emph{test} views.
}

\label{fig:gso_appearance_qual}
\end{figure*}

\begin{figure*}[t]
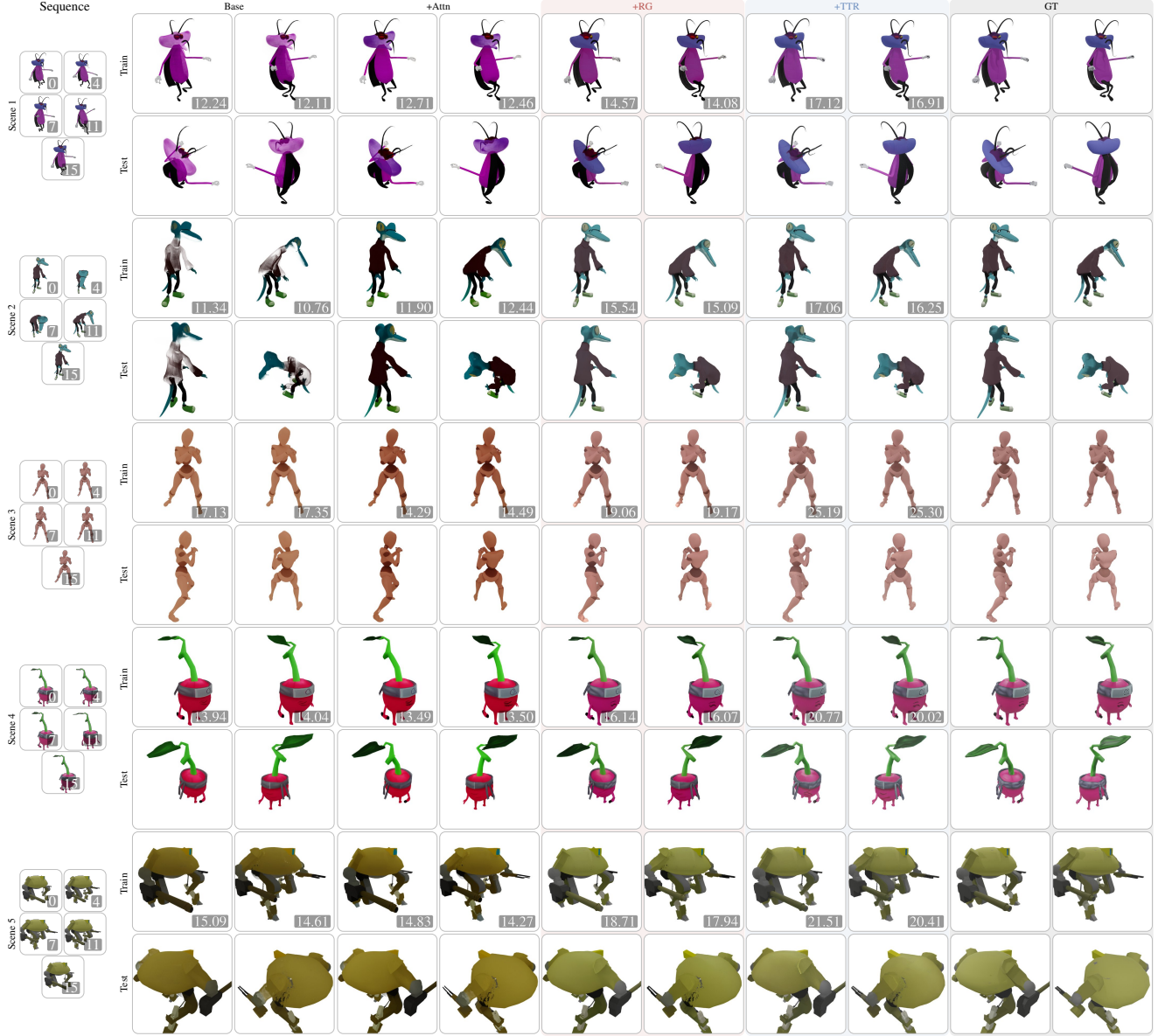

\centering
\bakedfig{qualitative_oab_appearance}

\caption{
\textbf{Qualitative appearance ablation on ActionBench with ground-truth geometry.}
Every column is our method on the per-frame GT mesh (shared GT shape and pose), so
only \emph{appearance} differs. 
Cumulative ladder, where each column adds one component to the column on its left.
The label names only what that step adds: Base (SAM3D appearance prediction + observation fusion) $\to$
+Attn (visibility attention bias) $\to$ +RG (rendering guidance) $\to$ +TTR
(test-time refinement). 
Each scene spans two subrows: rendered from
the input (\emph{train}) view and from the held-out (\emph{test}) view, each against its GT.
Each render is badged (bottom-right) with its per-scene appearance score: PSNR ($\uparrow$) on
the \emph{train} views and co-visible PSNR (co-PSNR, $\uparrow$;
\cref{sec:results:metrics}) on the \emph{test} views.
}

\label{fig:oab_appearance_qual}
\end{figure*}

\begin{figure*}[t]
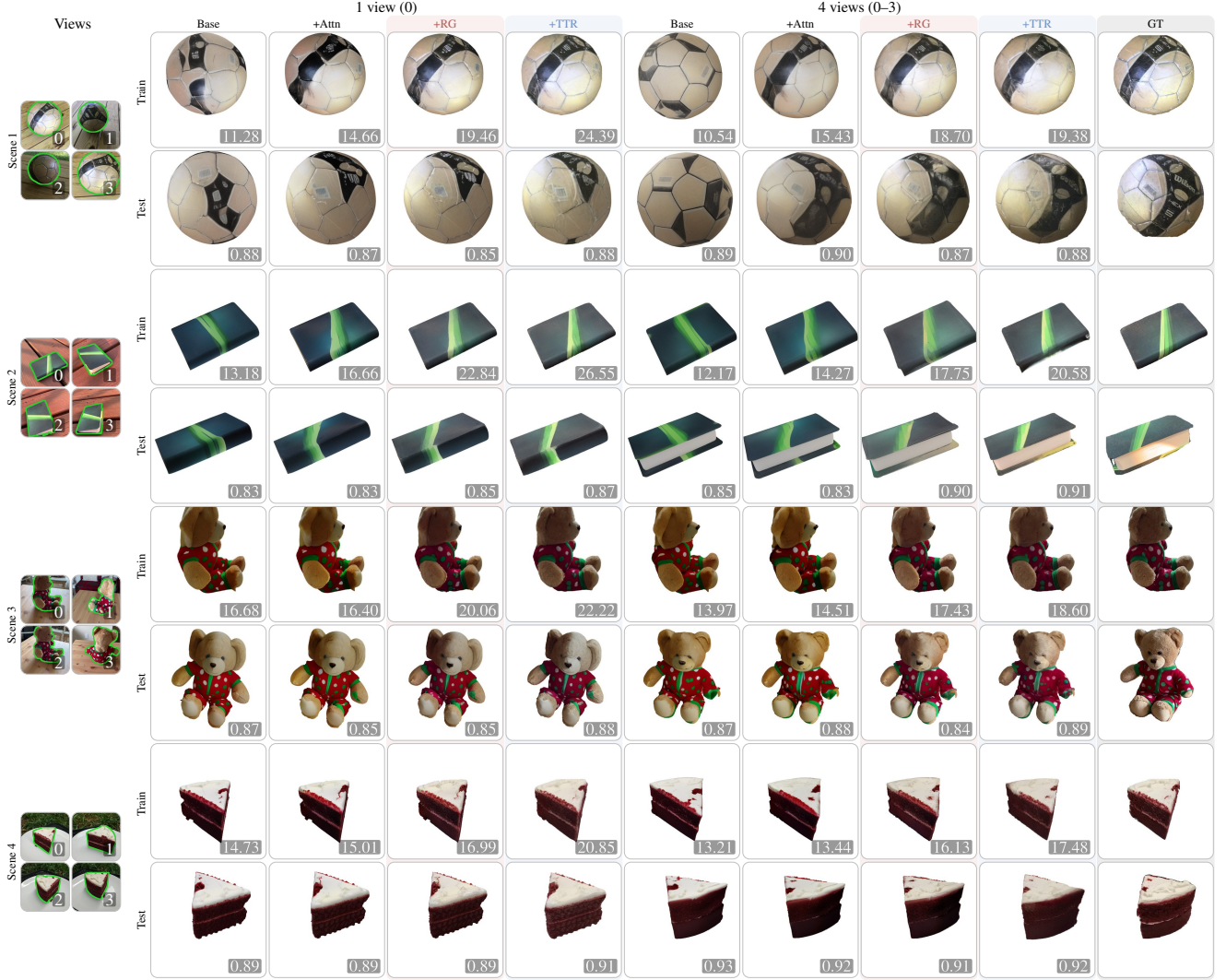

\centering
\bakedfig{qualitative_co3d_appearance}

\caption{
\textbf{Qualitative appearance ablation on CO3D with SAM3D geometry.}
Every column is our method on predicted shape and pose (CO3D has no ground-truth
mesh); SAM3D geometry from 1 input view and MV-SAM3D geometry from 4 input views. Only \emph{appearance} differs. 
Cumulative ladder, where each column adds one component to the column on its left.
The label names only what that step adds: Base (SAM3D appearance prediction + observation fusion) $\to$
+Attn (visibility attention bias) $\to$ +RG (rendering guidance) $\to$ +TTR
(test-time refinement). 
Each scene spans two subrows: the reconstruction rendered from
the first input (\emph{train}) view and from a held-out target (\emph{test}) view, each
against its GT. Each render is badged (bottom-right) with its per-scene appearance score:
PSNR ($\uparrow$) on the \emph{train} views and CLIP-I ($\uparrow$) on the \emph{test} views.
}

\label{fig:co3d_appearance_qual}
\end{figure*}

\begin{figure*}[t]
\centering
\bakedfig{qualitative_davis_appearance}

\caption{
\textbf{Qualitative appearance ablation on DAVIS with ActionMesh geometry.}
Every column is \method{} on the same ActionMesh-predicted temporal coarse shape, so only \emph{appearance} differs. 
Cumulative ladder, where each column adds one component to the column on its left.
The label names only what that step adds: Base (SAM3D appearance prediction + observation fusion) $\to$
+Attn (visibility attention bias) $\to$ +RG (rendering guidance) $\to$ +TTR
(test-time refinement).
Each scene spans two subrows: the
reconstruction rendered from the input (\emph{train}) view against its GT (the input frame
foreground-composited on white via the DAVIS annotation mask), and a synthesized
novel view at the same timestamp (\emph{test}).
Each render is badged with per-scene PSNR ($\uparrow$) on train views and CLIP-I ($\uparrow$) on test views.
}

\label{fig:davis_appearance_qual}
\end{figure*}

We cumulatively ablate the three components of our appearance stage on top of
SAM3D's appearance prediction (Base): visibility attention bias (Attn), rendering guidance (RG),
and test-time refinement (TTR).
Quantitative results are reported in
\cref{tab:appearance_static,tab:appearance_dynamic},
with qualitative comparisons in
\cref{fig:gso_appearance_qual,fig:oab_appearance_qual,fig:co3d_appearance_qual,fig:davis_appearance_qual}.

\Cref{tab:ablation_poses_full} reports cumulative pose ablations across static and dynamic settings.
All dynamic sequences are additionally evaluated for tracking against CoTracker3
(\cref{sec:supp:pseudo-tracks}).
Rotation averaging across frames leaves DAVIS essentially unchanged and modestly
reduces reprojection error and jitter on ActionBench.
Depth-grounded translation and scale then provide the largest
gains, reducing reprojection error by $1.4$--$2.8\times$ and relative jitter by
$2$--$4\times$ across the dynamic settings while also improving static-view PSNR.
ICP registration behaves differently across settings. For static reconstruction, it
halves Chamfer distance. On deforming geometry, its effect is dataset-dependent: on
DAVIS it improves masked depth but worsens appearance and reprojection error, likely
reflecting noise in the MapAnything depth and pose predictions; on ActionBench, where
geometry and cameras are more reliable, it also improves PSNR.

\subsection{Additional results}
\label{sec:supp:additional_results}

\begin{figure*}[t]
\centering
\bakedfig{qualitative_gso}

\caption{
\textbf{Qualitative comparison on GSO-30.}
Each render is badged (bottom-right) with its per-scene PSNR ($\uparrow$) on the
\emph{train} views and CLIP-I ($\uparrow$) on the \emph{test} views.
}

\label{fig:gso_comparison}
\end{figure*}

\begin{figure*}[t]
\centering
\bakedfig{qualitative_oab}

\caption{
\textbf{Qualitative comparison on ActionBench.}
Each render is badged (bottom-right) with its per-scene PSNR ($\uparrow$) on the
\emph{train} views and CLIP-I ($\uparrow$) on the \emph{test} views.
}

\label{fig:oab_comparison}
\end{figure*}

\begin{figure*}[t]
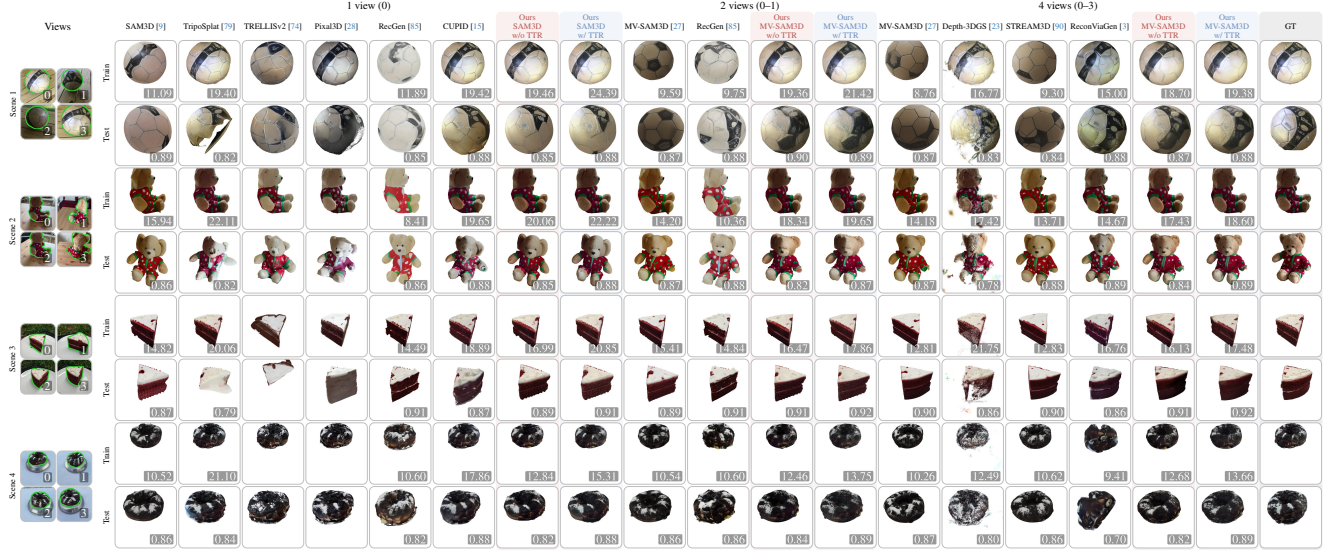

\centering
\bakedfig{qualitative_co3d}

\caption{
\textbf{Qualitative comparison on CO3D.}
Novel-view synthesis on 4 held-out CO3D (LaRa) scenes, reconstructed from 1, 2 and 4
input views. We compare each setting's baselines (SAM3D, TripoSplat, TRELLISv2,
Pixal3D, RecGen, and CUPID at 1 view; MV-SAM3D and RecGen at 2 views; MV-SAM3D, Depth-3DGS, STREAM3D, and ReconViaGen at 4 views) against our method on that setting's geometry backbone (SAM3D at 1 view, MV-SAM3D at 2 and 4
views). RecGen's released checkpoint takes at most two views, so we show it at its
maximum, in the 2-view group. 
TRELLISv2 and Pixal3D are qualitative-only
baselines: they predict mesh materials without environment lighting, so their renders
show unshaded base color (\cref{sec:results:baselines}). 
Each scene spans two subrows: the
reconstruction rendered from the first input (\emph{train}) view and from the held-out
target (\emph{test}) view, each against its GT. Each render is badged (bottom-right) with its
per-scene PSNR ($\uparrow$) on the \emph{train} views and CLIP-I ($\uparrow$) on the
\emph{test} views.
}

\label{fig:co3d_comparison}
\end{figure*}

\begin{figure*}[t]
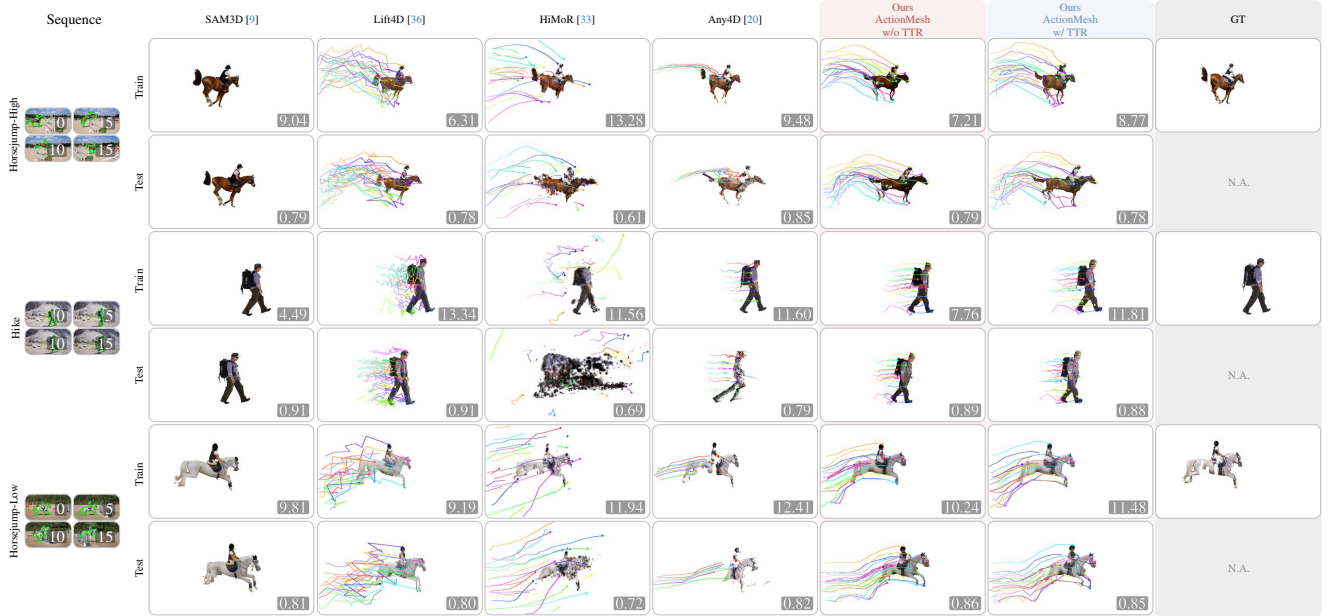

\centering
\bakedfig{qualitative_davis}

\caption{
\textbf{Qualitative comparison on DAVIS.}
Each scene spans two subrows. The \emph{train} subrow re-renders the input
viewpoint, against the input frame as GT. DAVIS is monocular and has no second camera, so
the \emph{test} subrow is instead a \emph{synthesized} novel view of the same timestamp.
Each reconstruction is badged (bottom-right) with its per-scene PSNR ($\uparrow$).
}

\label{fig:davis_comparison}
\end{figure*}

\begin{figure*}[t]
\centering
\bakedfig{qualitative_failures}

\caption{
\textbf{Failure cases.}
Dynamic geometry currently relies on an external predictor (ActionMesh) rather
than being directly grounded in the observations. Rotation stays prior-driven, corrected
only by noise-sensitive ICP registration. Both failure modes surface on dynamic scenes
below: incorrect world-space placement from registration, non-rigid deformation errors
inherited from ActionMesh, or a combination of the two, can strongly impact the final
reconstruction quality.
Each render is badged with per-scene PSNR ($\uparrow$) on \emph{train} views and
CLIP-I ($\uparrow$) on \emph{test} views (DAVIS shows a synthesized novel view in place
of a held-out test view; its GT test tile is left blank).
}
\label{fig:failure_cases}
\end{figure*}

\Cref{fig:gso_comparison,fig:oab_comparison,fig:co3d_comparison,fig:davis_comparison}
provide additional qualitative comparisons across all benchmarks.
\Cref{fig:failure_cases} shows two representative failure cases on dynamic
scenes.

\subsection{Dynamic shape injection}
\label{sec:supp:shape_injection}

For dynamic objects, we inject a sequence of fixed-topology meshes, where each vertex
indexes the same point across frames (as in ActionMesh and ActionBench), and invert
each deformed shape independently as described in \cref{sec:method:shape_injection}.
The shared topology provides both the deformation field used to animate the canonical
reconstruction and the coarse-geometry correspondences used to fuse per-frame
appearance.

We recover a per-vertex rigid deformation directly from the mesh sequence. Let
$\verts_n^{(k)}$ denote the position of vertex $n$ at frame $k$, and
$\bar{\verts}_n=\verts_n^{(c)}$ its position in the canonical frame $c$. For each
canonical vertex $n$ and frame $k$, we estimate a local rotation
$R_n^{(k)}\in\SO(3)$ by aligning its canonical neighborhood $\mathcal{K}(n)$,
the $K=8$ nearest canonical vertices, to its deformed counterpart using Kabsch:
\[
R_n^{(k)}
=
\argmin_{R\in\SO(3)}
\sum_{w\in\mathcal{K}(n)}
\left\|
R(\bar{\verts}_w-\bar{\verts}_n)
-
(\verts_w^{(k)}-\verts_n^{(k)})
\right\|^2 .
\]
Together with the deformed vertex positions $\verts_n^{(k)}$, these rotations define
the deformation field $\defm_{c\rightarrow k}$ of \cref{sec:method:shape_injection},
which maps canonical vertex $n$ to $(\verts_n^{(k)},R_n^{(k)})$ and is the identity at
the canonical frame. We use $\defm$ to warp the
canonical Gaussians, with their centers following the vertex motion and their
covariances following the local rotations.

\begin{figure}[t]
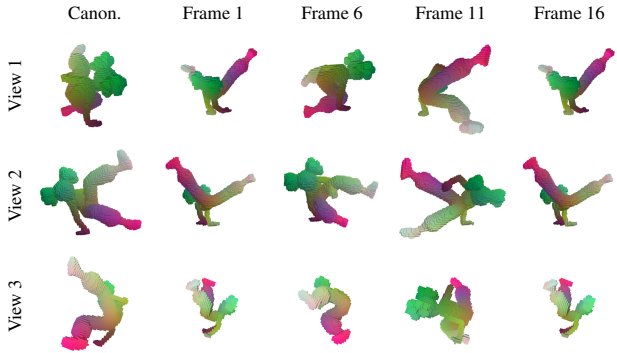

\centering
\bakedfig{voxel_correspondence}

\caption{%
\textbf{Per-frame to canonical voxel correspondence} on ActionBench
(scene \texttt{000-003}).
Each frame's mesh is voxelized in its own bounding box (its native per-frame
grid); every voxel is matched to a canonical voxel through the shared fixed topology
(\cref{sec:supp:shape_injection}) and colored by that canonical voxel's position, so
a correct correspondence keeps each body part a constant color throughout the
sequence. In this example, columns show the canonical frame $c{=}5$ and frames $1, 6, 11, 16$; rows show three
orbit viewpoints.
}
\label{fig:voxel_correspondence}
\end{figure}

Fixed topology provides correspondences between each frame's voxel grid and the
canonical grid. Each per-frame voxel is thus associated with a canonical voxel, with
the correspondence reducing to the identity at the canonical frame
(\cref{fig:voxel_correspondence}).

\paragraph{Per-frame appearance fusion.}
At every ODE step, we use these correspondences to map each frame's prediction
onto the canonical grid, where the per-frame predictions are fused.
The voxel correspondence can be many-to-one when multiple native voxels map to the same canonical cell.
Collisions within each canonical voxel are masked using visibility and then averaged, as in visibility-weighted fusion (\cref{sec:method:appearance}).

\paragraph{Pose denoising with clean shape latents.}
When the shape latents are injected, we fix the clean shape latent $\lat^{\voxels}_{1}$ and predict the object's pose.
Because SAM3D jointly denoises shape and pose, the network expects a shape token at
every ODE step $\ftime$. 
We fix the target geometry and preserve the
training-time trajectory using an interpolated intermediate shape token $\lat^{\voxels}_{\ftime} = \ftime\,\lat^{\voxels}_{1} + (1-\ftime)\,\lat^{\voxels}_{0}$, 
where $\lat^{\voxels}_{0}\sim\Normal(\mathbf{0},\eye)$ is fixed.

\subsection{Voxel-to-patch mapping}
\label{sec:supp:visibility-details}
For each tokenized conditioning stream, we map a voxel's pixel footprint (\cref{sec:visibility}) through the corresponding crop--pad--resize transformation (mask-bounding-box crop for cropped streams; pad-to-square for full-image streams) and bin it into the $37{\times}37$ DINOv2 patch grid. The resulting patch set $\patchset_i(b)$ is dilated by one patch to account for projection quantization.

During visibility-weighted fusion (\cref{sec:method:appearance}), a canonical voxel that no voxel of frame $b$ maps to receives no contribution from that frame; every canonical voxel is reached at least by the canonical frame, whose correspondence is the identity.

\subsection{Passive-stream compensation}
\label{sec:supp:passive-compensation}
The visibility attention bias (\cref{sec:method:appearance}) increases the logits of
projected patches in the active RGB conditioning groups, while softmax is normalized
over all conditioning tokens. This unintentionally reduces the attention assigned to
unmodified mask and global tokens.

\begin{figure*}[t]
\centering
\includegraphics[width=\textwidth]{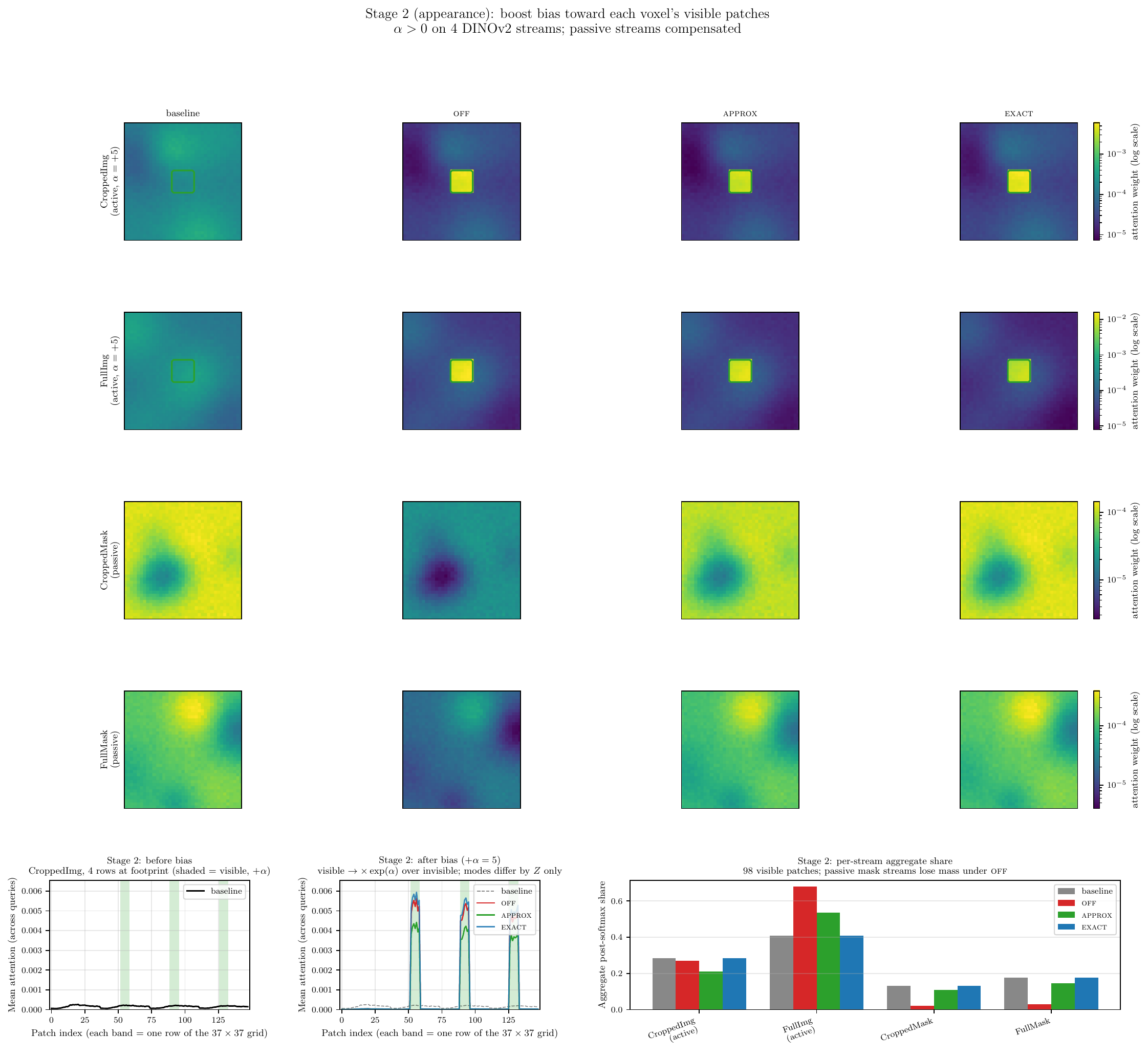}
\caption{%
\textbf{Simulated effect of passive-stream compensation for visibility-biased attention.} We visualize four simulated DINOv2 $37{\times}37$ streams. Visibility bias ($\alpha>0$) boosts projected patches in the active RGB streams (CroppedImg, FullImg), which in turn suppresses the passive mask streams (CroppedMask, FullMask) through softmax normalization. Without compensation (\textsc{off}), passive-stream attention decreases; the closed-form approximation (\textsc{approx}) nearly restores its unbiased level and closely matches the post-softmax oracle (\textsc{exact}).
}
\label{fig:vis_attn_comp_s2}
\end{figure*}

We compensate with a per-voxel offset. Let $\kappa$ denote the fraction of active RGB
tokens receiving bias $\alpha$. Assuming uniform pre-bias logits within the active
tokens, their pooled exponentiated mass increases by
$\kappa e^\alpha + 1-\kappa$. We therefore subtract
\begin{equation}
\beta
=
\log\!\left(\kappa e^\alpha + 1-\kappa\right)
\label{eq:passive-compensation}
\end{equation}
from all active RGB logits. Under this approximation, the pooled active mass is restored,
and hence the baseline attention share of mask and global tokens is preserved
(\cref{fig:vis_attn_comp_s2}).
Independently of this assumption, the common offset leaves within-active ratios unchanged,
so visible patches retain their exact $e^\alpha$ multiplicative boost. In practice,
$\beta$ is computed per voxel from the active bias row and folded into the additive bias.

We also evaluate a post-softmax oracle that uses the actual attention distribution to
restore the baseline mass of each active conditioning group individually. Unlike the
closed-form approximation, this accounts for non-uniform logits. On GSO-30 and CO3D, no compensation (\textsc{off}), the closed-form approximation
(\textsc{approx}), and the oracle (\textsc{exact}) yield nearly identical reconstruction metrics,
indicating that the compensation is primarily a principled correction rather than a
source of measurable quality gains. We use \textsc{approx} because it adds essentially
no cost, whereas \textsc{exact} incurs a $1.3$--$1.8\times$ overhead in the appearance
block.

\subsection{Rendering loss details}
\label{sec:supp:render-loss}
We detail the terms of the per-observation rendering loss
\eqref{eq:render-loss-terms}, used in 
appearance prediction (\cref{sec:method:appearance}) and in test-time refinement (\cref{sec:method:ttt}). 
Here, we drop the observation
index $b$: every quantity below belongs to one observation, rendered under its fixed
pose.
Since only the object is rendered, the observed image $\img$ enters each term
as the composite
$\bar\img=\mask\odot\img+(1-\mask)\eta$, where the renderer background color
$\eta$ is resampled uniformly at each
step to prevent the model from encoding a fixed background into the appearance.
Each term is as follows:
\begin{equation}
\begin{aligned}
\lossrgb &= \lVert \hat\img - \bar\img \rVert_1, \\
\lossssim &= \bigl(1 - \operatorname{SSIM}(\hat\img, \bar\img)\bigr), \\
\losslpips &= \operatorname{LPIPS}(\hat\img, \bar\img), \\
\lossdepth &= \frac{\lVert \mask \odot (\hat\depth - \depth) \rVert_1}
{\lVert \mask \rVert_1} .
\end{aligned}
\label{eq:render-loss-defs}
\end{equation}
$\lossrgb$ is evaluated over an image pyramid at scales
$(1,\tfrac12,\tfrac14)$ with weights $(0.5,0.3,0.2)$, while $\losslpips$ is computed
on the object's bounding-box crop. $\lossdepth$ is the object-masked mean absolute
error in $z$-depth.

\paragraph{Gradient masking.}
Generated Gaussians are semi-transparent; therefore, primitives decoded from voxels that are not directly observed still contribute marginally to the rendered image. Consequently, the loss produces gradients for regions of the object that it cannot directly constrain.
Left unchecked, these push the unseen surface away from the generative prior,
producing holes and floaters in novel views. 
We apply a
\emph{per-Gaussian stop-gradient}: Gaussians occluded in one observation are
detached before rendering that observation, so the rendering gradient flows only through directly observed geometry.

\paragraph{Opacity anchor.}
During test-time refinement, the adapted decoder may reduce the opacity of occluded Gaussians to lower the reconstruction error, effectively hollowing out unseen surfaces. We therefore introduce an \emph{opacity anchor} that penalizes deviations from the initial opacity values. We find that this simple constraint is sufficient to preserve solid object geometry in novel views.
The anchor is one-sided, penalizing only opacity drops below the initial value
$\sigma_j^0$ over the set $\mathcal{O}$ of Gaussians occluded in every observation,
\begin{equation}
\loss_{\mathrm{op}} = \lw_{\mathrm{op}}\,\frac{1}{|\mathcal{O}|}\sum_{j\in\mathcal{O}}
\max\bigl(0,\,\sigma_j^0-\sigma_j\bigr)^2,
\qquad \lw_{\mathrm{op}}=0.1 .
\end{equation}

\subsection{Test-time refinement details}
\label{sec:supp:ttr}
Test-time refinement (\cref{sec:method:ttt}) optimizes the canonical latent $\slat$,
Gaussian-decoder adapter $\Delta\decGauss$, and pose $\layout$ for $100$ iterations per
object using AdamW with PyTorch-default $\beta$, $\epsilon$, and weight decay. Each
iteration renders one randomly sampled observation (a stochastic estimate
of the sum over observations in \eqref{eq:ttt}), and objects in multi-object scenes are refined
sequentially. Learning rates are linearly warmed up for $10$ iterations and then fixed
at $10^{-4}$ for $\slat$, $10^{-3}$ for $\Delta\decGauss$, and
$10^{-2}/10^{-3}$ for pose rotation/translation. Scale is not refined.

$\Delta\decGauss$ is a rank-$4$ DoRA~\cite{liu2024dora} adapter applied to every linear
layer of the Gaussian decoder, with $\alpha=0.5$ and rank-stabilized scaling
$\alpha/\sqrt{r}$~\cite{kalajdzievski2023rank}. Its second factor is zero-initialized,
so refinement starts exactly from the unadapted decoder prediction.

The latent anchor in \eqref{eq:ttt} is an MSE to the initial latent $\slat^0$ with
$\anchorw=0.01$. The rendering objective follows \eqref{eq:render-loss-terms} with the
same weights. Visibility for its stop-gradient mask and opacity anchor is determined by
a depth test against rendered expected depth with relative behind-surface tolerance
$0.02$.

All remaining parameters, including the SAM3D model and mesh decoder, remain frozen.
The in-loop Gaussian-decoder forward uses \texttt{bfloat16} autocast to reduce memory.

\subsection{Depth-3DGS baseline}
\label{sec:supp:depth3dgs}

Depth-3DGS is a prior-free, per-scene 3DGS~\cite{kerbl3Dgaussians} baseline using the
same cameras and depth as \method{}. We initialize up to $100$k Gaussians by
unprojecting foreground depth pixels into a common frame.
We optimize the standard 3DGS photometric loss ($L_1$ and SSIM with weight $0.2$)
over the masked foreground, together with an inverse-depth $L_1$ loss weighted by
$10^{-2}$.
Each scene is optimized for $1000$ iterations, with adaptive density control rescaled
from the standard $30$k-iteration schedule and opacity resets disabled. We use
degree-$0$ spherical harmonics for view-independent color.

\subsection{ICP registration}
\label{sec:supp:registration}
The ICP registration step of \cref{sec:method:poses_init} aligns the coarse shape to the
metric input pointmap using one similarity transform per object, which is composed with
the pose of every frame. This absorbs systematic alignment error while preserving
per-frame motion.
We match observed pointmap samples to points rendered from the coarse shape rather than
using a pixel-aligned loss, which under small misregistration can compare unrelated
surfaces. Matching from observed to rendered points also naturally handles occlusion,
since unobserved parts of the object are not required to explain any measurements.
Let $\mathcal{P}$ be the observed foreground pointmap and
$\hat{\mathcal{P}}(\layout)$ the surface unprojected from the depth rendered under the
current pose. We minimize the one-directional nearest-neighbor objective
\begin{equation}
\loss_{\mathrm{reg}}(\layout)
= \frac{1}{|\mathcal{P}|}\sum_{p\in\mathcal{P}}
  \min_{q\in\hat{\mathcal{P}}(\layout)}\lVert p-q\rVert^2
\end{equation}
by gradient descent over the similarity transform.

\subsection{Pose refinement for baselines}
\label{sec:supp:baseline-poses}
Pose misregistration can dominate pixel-aligned
object reconstruction metrics. We therefore report pose-refined results, keeping shape and appearance fixed
while optimizing only rotation, translation, and scale against the input observations.
For pose-predicting methods (e.g.\ SAM3D and MV-SAM3D), refinement starts from the
predicted pose. For methods without pose prediction, such as TripoSplat and TRELLISv2,
we initialize translation and scale as in \method{}, obtain rotation by rendering-based
search, and apply the same refinement.
This isolates reconstruction quality from pose-estimation error while giving each method
its best alignment to the inputs. Methods that reconstruct directly in the input camera
frames are exempt, as they are already aligned.

\subsection{Pseudo-ground-truth 2D tracking}
\label{sec:supp:pseudo-tracks}

We use the off-the-shelf point tracker CoTracker3~\cite{karaev2024cotracker3} as a
pseudo-reference. For each sequence, we place a uniform
grid of queries on the first-frame object mask and track them throughout the clip.
To reduce tracker errors, we retain only cycle-consistent queries whose backward track
returns within $0.3$ pixels of their starting point. This removes $40$--$60\%$ of queries and
more than halves the reference error on annotated sequences.

For reconstruction-based methods, each query is matched to the reconstructed surface
point whose projection is nearest in the first frame and propagated through the recovered
pose and deformation, with visibility determined by geometric self-occlusion. The
initial query-to-surface distance bounds the attainable tracking error,
remaining below one pixel for every method on both datasets. Since the same
CoTracker3 tracks are used for every method, tracking measures the temporal consistency
of the recovered 4D reconstruction.

Beyond the DAVIS $\delta_{\mathrm{avg}}$ and reprojection error reported in
\cref{tab:dynamic_full}, the reference's occlusion labels yield an average Jaccard of
$0.42$ for HiMoR against $0.37$ for \method{} (with or without test-time refinement)
and $0.27$ for Lift4D; Any4D exports no point visibility, so AJ is unavailable for it.
On ActionBench the same protocol places Any4D closest to CoTracker3
($\delta_{\mathrm{avg}}{=}0.71$), ahead of HiMoR ($0.65$) and \method{}
($0.50$, with or without test-time refinement).
These metrics measure agreement with CoTracker3 rather than absolute tracking accuracy
and should therefore be interpreted as a coarse relative comparison, with which
a pure point tracker like Any4D naturally agrees most closely.

\subsection{Datasets and scenes}
\label{sec:supp:datasets}
\emph{GSO-30.} We use the 30 EscherNet~\cite{kong2024eschernet} objects
(\scene{alarm}, \scene{backpack}, \scene{bell}, \scene{blocks}, \scene{chicken},
\scene{cream}, \scene{elephant}, \scene{grandfather}, \scene{grandmother}, \scene{hat},
\scene{leather}, \scene{lion}, \scene{lunch_bag}, \scene{mario}, \scene{oil},
\scene{school_bus1}, \scene{school_bus2}, \scene{shoe}, \scene{shoe1}, \scene{shoe2},
\scene{shoe3}, \scene{soap}, \scene{sofa}, \scene{sorter}, \scene{sorting_board},
\scene{stucking_cups}, \scene{teapot}, \scene{toaster}, \scene{train}, \scene{turtle}),
each rendered from 25 views at $512^2$. Views 0--9 are input candidates
(the first \mbox{\#V} are used; view 0 for monocular input), while views 10--24 are held out.

\emph{CO3D.} We evaluate ten CO3Dv2~\cite{reizenstein21co3d} sequences
(\scene{34_1479_4753}, \scene{50_2928_8645}, \scene{123_14363_28981},
\scene{167_18184_34441}, \scene{189_20393_38136}, \scene{247_26441_50907},
\scene{247_26469_51778}, \scene{374_42274_84517}, \scene{391_47032_93657},
\scene{415_57121_110109}) following LaRa~\cite{LaRa2024ECCV}.
Camera poses are clustered into four groups; one view per cluster forms the input set
(the first \mbox{\#V} are used), and four further views, chosen greedily to maximize
angular separation from the input views and from each other, are held out for testing.
Foreground masks are provided by CO3D.

\emph{ActionBench.} We use six ActionBench~\cite{ActionMesh2026} scenes
(\scene{000-001_09}, \scene{000-004_28}, \scene{000-004_59}, \scene{000-007_17},
\scene{000-007_84}, \scene{000-008_95}), re-rendered from the source Sketchfab
animations with known cameras. We render 16 keyframes from a fixed input camera and
export the deformed mesh per frame; an additional upper-hemisphere camera at each
timestamp provides the held-out view.

\emph{DAVIS.} We use ten sequences
(\scene{bear}, \scene{blackswan}, \scene{camel}, \scene{car-roundabout}, \scene{cows},
\scene{hike}, \scene{horsejump-high}, \scene{horsejump-low}, \scene{mallard-water},
\scene{rhino}), sampling 16 frames at stride 2 from the first 32.
DAVIS annotations provide object masks.

GSO-30 and ActionBench use ground-truth depth and camera poses; CO3D and DAVIS use
MapAnything~\cite{keetha2026mapanything} predictions from the input observations.

\subsection{Metric reliability}
\label{sec:supp:metrics}
Overall, the quantitative rankings broadly agree with the qualitative comparisons
(\cref{fig:gso_comparison,fig:oab_comparison,fig:co3d_comparison,fig:davis_comparison}),
but each metric has limitations. PSNR remains sensitive to small residual
misalignments after pose refinement, so we interpret it together with LPIPS, which is
more tolerant to local shifts, and co-PSNR, which evaluates only regions constrained by
an input view.
CLIP-I varies over a relatively narrow range ($0.71$--$0.91$ in
\cref{tab:full_benchmarks}) and is comparatively tolerant to pose because each image is
cropped to its own object bounding box. 
On ActionBench, per-frame SAM3D leads CLIP-I
(\cref{tab:dynamic_full}), whereas our reconstruction inherits ActionMesh deformation
and placement errors. With ground-truth geometry, the same appearance stage reaches
$0.92$ (\cref{tab:appearance_dynamic}) versus $0.86$, indicating that geometry quality
strongly affects the score. TTR can further reduce CLIP-I when it fits the input view to
geometry that is inaccurate from the held-out viewpoint.
On DAVIS, CLIP-I compares each synthesized novel view with its corresponding input
frame, measuring appearance consistency under viewpoint change rather than
absolute reconstruction accuracy.

\end{document}